%% file: neurips_2024.tex
\documentclass{article}

\usepackage[final, nonatbib]{neurips_2024}

\usepackage[utf8]{inputenc} % allow utf-8 input
\usepackage[T1]{fontenc}    % use 8-bit T1 fonts
\usepackage{hyperref}       % hyperlinks
\usepackage{url}            % simple URL typesetting
\usepackage{booktabs}       % professional-quality tables
\usepackage{amsfonts}       % blackboard math symbols
\usepackage{nicefrac}       % compact symbols for 1/2, etc.
\usepackage{microtype}      % microtypography
\usepackage{xcolor}         % colors
\usepackage{graphicx} 
\usepackage{amsmath}
\usepackage{amssymb}
\definecolor{ForestGreen}{RGB}{34,139,34}
\usepackage{wrapfig,lipsum,booktabs}
\usepackage{multirow}
\usepackage{bm}
\usepackage{amsthm}
\usepackage{wrapfig,lipsum,booktabs}
\usepackage{multirow}
\usepackage{amsthm}
\usepackage{thmtools,thm-restate}
\usepackage{algpseudocode}
\usepackage{algorithm}
\newtheorem{theorem}{Theorem}[section]
\newtheorem{lemma}[theorem]{Lemma}
\usepackage{pythonhighlight}

\title{Sparse High Rank Adapters}

\author{\hspace{-1mm}Kartikeya Bhardwaj\thanks{Equal contribution. $^\dag$Work done while employed at Qualcomm AI Research. $^\ddag$Qualcomm AI Research is an initiative of Qualcomm Technologies, Inc. $^\P$Code: \url{https://github.com/Qualcomm-AI-research/SHiRA}.} \ $^\S$\enspace Nilesh Prasad Pandey$^*$$^\dag$\enspace Sweta Priyadarshi$^\dag$\enspace Viswanath Ganapathy$^\dag$ \\
\textbf{Shreya Kadambi\quad Rafael Esteves\quad Shubhankar Borse\quad Paul Whatmough$^\S$ } 
\\
\textbf{Risheek Garrepalli\quad Mart Van Baalen\quad Harris Teague$^\S$\quad Markus Nagel$^\S$}  \\
Qualcomm AI Research$^\ddag$ \\
\texttt{$^\S$\{kbhardwa,pwhatmou,hteague,markusn\}@qti.qualcomm.com}%\\
}

\begin{document}

\maketitle

\vspace{-3mm}
\begin{abstract}\vspace{-4mm}
Low Rank Adaptation (LoRA) has gained massive attention in the recent generative AI research. One of the main advantages of LoRA is its ability to be fused with  pretrained models, adding no overhead during inference. However, from a mobile deployment standpoint, we can either avoid inference overhead in the fused mode but lose the ability to switch adapters rapidly, or suffer significant (up to 30\% higher) inference latency while enabling rapid switching in the unfused mode. LoRA also exhibits concept-loss when multiple adapters are used concurrently. In this paper, we propose Sparse High Rank Adapters (SHiRA), a new paradigm which incurs no inference overhead, enables rapid switching, and significantly reduces concept-loss. Specifically, SHiRA can be trained by directly tuning only $1$-$2\%$ of the base model weights while leaving others unchanged. This results in a highly sparse adapter which can be switched directly in the fused mode. We further provide theoretical and empirical insights on how high sparsity in SHiRA can aid multi-adapter fusion by reducing concept loss. Our extensive experiments on LVMs and LLMs demonstrate that finetuning only a small fraction of the parameters in the base model significantly outperforms LoRA while enabling both rapid switching and multi-adapter fusion. Finally, we provide a latency- and memory-efficient SHiRA implementation based on Parameter-Efficient Finetuning (PEFT) Library which trains at nearly the same speed as LoRA while consuming up to $16\%$ lower peak GPU memory, thus making SHiRA easy to adopt for practical use cases. To demonstrate rapid switching benefits during inference, we show that loading SHiRA on a base model can be $5\times$-$16\times$ faster than LoRA fusion on a CPU.$^\P$
\end{abstract}\vspace{-1mm}
% At inference time, we also demonstrate rapid switching benefits of SHiRA on a CPU: loading SHiRA on top of the base model can be 5×-16× faster than LoRA fusion.

\input{sections/01_introduction}\label{sec:introduction}

\input{sections/02_related_work}\label{sec:Related_Work}
\input{sections/03_method}\label{sec:method}
\input{sections/06_theory}\label{sec:theory}
\input{sections/04_experiments}\label{sec:experiments}
\input{sections/05_conclusion}\label{sec:conclusion}

\bibliographystyle{plain}
\bibliography{neurips} 

\newpage

\appendix

\input{sections/07_appendix}\label{sec:appendix}

\clearpage

\input{sections/checklist}
\end{document}

%% file: sections/01_introduction.tex
\vspace{-0.3cm}
\section{Introduction}\label{sec::intro}
\vspace{-0.3cm}
Low Rank Adaptation (LoRA)~\cite{lora} is an established technique to tune the behavior of large generative models such as Large Language Models (LLMs)~\cite{llama2, llama} and Stable Diffusion~\cite{sd, sdxl}. As the name suggests, LoRA requires very few parameters since it trains low rank projection weights that consume very low memory during the finetuning process while producing excellent results. Moreover, these low rank weights can be fused analytically into the base model, thereby incurring no additional overhead during inference.

Despite its success, there are still several limitations of low rank adaptation methods. First, if LoRA parameters are fused into the corresponding pretrained base model weights, they modify the entire weight tensor. Therefore, deploying LoRA on large models such as LLaMA-1/2 (7B+ parameters) or Stable Diffusion (1.5B+ parameters) on mobile devices would require changing a large number of weights during inference. Consequently, for mobile scenarios, if an application requires \textit{rapid adapter switching}, existing low rank methods would incur a significant memory and latency cost. This is a major deployment challenge because, unlike large GPUs, local memory of small AI accelerators is limited and cannot store all weights at the same time. These challenges can be partially addressed by running LoRA in unfused mode; however, unfused inference can incur as high as $\bm{30\%}$ \textbf{additional latency} compared to the base model~\cite{hflora} (see section~\ref{sec::approachMobile} for details). This increased inference time in unfused mode and time for adapter switching significantly hampers user experience; hence, this is an important problem which has been a focus of recent research by various industries~\cite{gunter2024apple}.
Second, LoRA has a well-known limitation called \textit{concept loss} when using multiple concurrent adapters, e.g., combining multiple style transfer adapters, etc. Specifically, it has been well documented~\cite{yu2023language, shah2023ziplora, gu2024mix} that a simple additive merging of multiple LoRA adapters leads to concept loss of one or more adapters. Finally, recent literature also contributes important theoretical and empirical knowledge towards the value of \textit{high rank adapters}. For instance, Kalajdzievski~\cite{rslora} shows that the high rank adapters can greatly outperform low rank adapters when used with correct scaling factors. This calls for further investigation into whether other high rank adapters would significantly outperform LoRA.
\begin{figure}[t]
  \centering
   \includegraphics[width=0.85\linewidth]{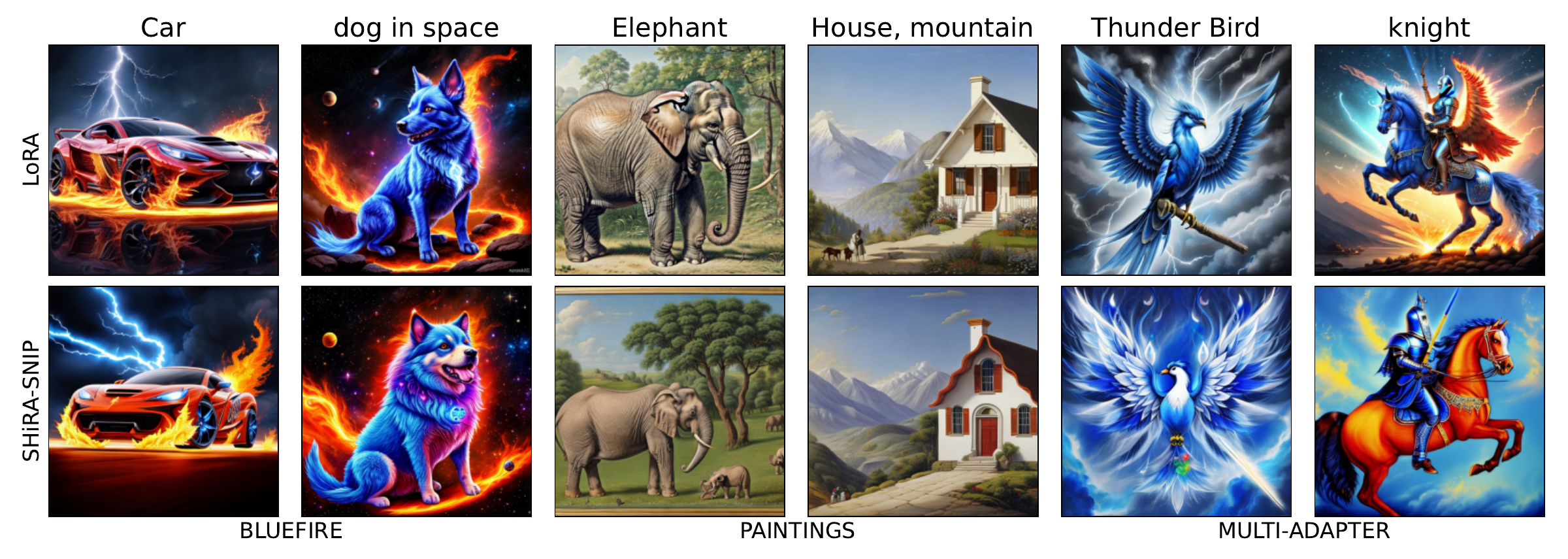}\vspace{-3mm}
   \caption{\underline{S}parse \underline{Hi}gh \underline{R}ank \underline{A}dapters (SHiRA): Changing about $1$-$2\%$ weights of the pretrained generative model is often sufficient to achieve high performance. Due to its extreme sparsity, SHiRA enables rapid switching and also reduced concept loss during multi-adapter fusion. In contrast, LoRA modifies majority of parameters when fused, thus prohibiting rapid switching on mobile devices, and also experiences concept loss during multi-adapter fusion. For LoRA, elephant for single ``paintings'' adapter case has artifacts (extra/broken tusks); bird and knight for multi-adapter case lose ``paintings'' concept and keep only the ``blue fire'' effects. SHiRA does not experience these issues.\vspace{-3mm}}
   \label{fig:introFigure}
\end{figure}

In view of the above, we address the following \textbf{key problems} in this paper: (\textit{i})~How can we perform rapid switching for fused adapters? (\textit{ii})~Is there a simpler solution for multi-adapter fusion to reduce concept loss? (\textit{iii})~Can we build high rank adapters that have high expressive power without significantly increasing the training or inference costs?

To this end, we propose \underline{S}parse \underline{Hi}gh \underline{R}ank \underline{A}dapters (SHiRA), a single solution to all three problems above. SHiRA is a highly sparse but a high rank adapter which relies on training only a very small subset of parameters from the original pretrained network. One of the crucial insights we demonstrate is that even finetuning merely $1$-$2\%$ parameters of the pretrained generative model is sufficient to achieve high performance on many adapter tasks (see Fig.~\ref{fig:introFigure}). 
However, unlike LoRA layers that modify all parameters in the weight tensors in the fused mode, SHiRA still keeps a very low percentage of parameters that need to be switched, thus enabling rapid switching at inference time. Moreover, since the pretrained weights are huge, SHiRA being a very sparse adapter greatly aids multi-adapter fusion by significantly reducing concept loss. Finally, we theoretically and emprically analyze the high rank vs. sparsity properties of SHiRA and why that helps with adapter performance. 

Overall, we make the following \textbf{key contributions}:
\vspace{-0.1cm}
\begin{itemize}\setlength{\itemsep}{-0.2em}
    \item We propose SHiRA, a new high rank adapter paradigm to demonstrate that changing as few as $1$-$2\%$ parameters of the original network is sufficient for adaptation. Our crucial insight is that even the most basic masking criteria (to identify the top $1$-$2\%$ parameters) enable SHiRA to significantly outperform LoRA on diverse vision and language tasks.
    \item SHiRA enables on-device \textit{rapid adapter switching} and provides a natural multi-adapter fusion technique due to high sparsity, thus, significantly reducing \textit{concept loss}. We also theoretically analyze SHiRA through the lens of \textit{high rank adaptation} vs. sparsity.
    \item We conduct extensive experiments on LLMs (LLaMA-7B, LLaMAv2-7B) and LVMs (Stable Diffusion, SDXL) where we demonstrate that SHiRA significantly outperforms LoRA on both single- and multi-adapter tasks. On LLMs, we show that SHiRA achieves up to $2.7\%$ better accuracy than LoRA on commonsense reasoning. SHiRA also complements advanced variants of LoRA such as DoRA~\cite{dora} and can be easily applied on top of them.
    \item Finally, on the training side, we provide a PEFT-based latency- and memory-efficient implementation for SHiRA which trains nearly as fast as standard LoRA while consuming $16\%$ lower peak GPU memory. Beyond PEFT, we provide a simple way to turn any trainer into SHiRA finetuning. For inference, we demonstrate that SHiRA weights can be loaded on a CPU up to $5\times$-$16\times$ faster than equivalent LoRA fusing, thereby enabling rapid switching.
\end{itemize}

The rest of this paper is organized as follows: section~\ref{sec::rel} presents the background and related work. We propose SHiRA in section~\ref{sec::approach} while describing its theoretical properties in section~\ref{sec::theory}. We then conduct extensive experiments for SHiRA in section~\ref{sec::exp}. Finally, we discuss the key findings in section~\ref{sec::discussion} and conclude the paper in section~\ref{sec::conclusion}.
\vspace{-0.3cm}

%% file: sections/02_related_work.tex
\section{Background and Related Work}\label{sec::rel}
\vspace{-0.3cm}
\subsection{Background: Edge Deployment Challenges for LoRA}\label{sec::approachMobile}
\vspace{-0.2cm}
There are three existing deployment options for LoRA: (\textit{i})~fuse the adapter offline and then deploy on-device: this 
changes a large fraction of the weight tensors compared to base model which prohibits rapid switching since it will increase DRAM traffic considerably; (\textit{ii})~keep the adapter unfused and run the inference in unfused mode: this can help with rapid switching but would incur  significant additional (up to 30\% higher) latency as shown in~\cite{hflora} since we would have LoRA branches in the forward pass during inference; (\textit{iii})~use the Huggingface/Diffusers pipeline~\cite{hflora} (built for server-grade GPUs) for mobile inference. This pipeline consists of \texttt{load}$\rightarrow$\texttt{fuse}$\rightarrow$\texttt{inference}$\rightarrow$\texttt{unfuse}$\rightarrow$\texttt{unload} to switch adapters. Here, unfused LoRA-A and LoRA-B weights (see Fig.~\ref{fig:shiraApproach}(a)) are first loaded into the memory and then fused into the base model by computing $W_{new} = W+AB$; this new weight is used for inference. To switch the adapter, we can unfuse the adapter as $W=W_{new}-AB$ and then unload existing LoRA weights to load the new ones. We provide further evidence in Appendix~\ref{sec::approachMobileContd} to  demonstrate that such a pipeline is not feasible for edge devices. This is primarily because edge devices are memory-limited and not all weights of large generative models can be stored in the local memory at the same time. Hence, loading and fusing needs to happen layerwise on a mobile device that obviously results in massive inference latency costs.
\vspace{-0.2cm}
\subsection{Related Work}
\vspace{-0.15cm}
%Our work is most related to the highly popular LoRA, its variants, sparse adapters, partial finetuning, and multi-adapter fusion techniques, as discussed below.

\textbf{LoRA, its variants, and sparse adapters.} Many LoRA variants exist in literature: DoRA~\cite{dora}, LoRA+~\cite{hayou2024lora+}, VeRA~\cite{kopiczko2023vera}, LoRA-FA~\cite{zhang2023lora}, RS-LoRA~\cite{rslora}, among many others. The crucial difference between this literature and our work is that we develop a high rank adapter without increasing training and inference costs. Also, for such methods, the final fused adapter still updates all elements in the pretrained weight tensor, thus prohibiting rapid switching. Moreover, for completeness, we will also show that SHiRA is orthogonal to and can be applied on top of some of the latest, more advanced LoRA variants such as DoRA~\cite{dora}  while preserving the benefits of rapid switching.

A few other LoRA variants have also explored a combination of sparsity and low rank adaptation. Examples include RoSA~\cite{nikdan2024rosa}, SoRA~\cite{ding2023sparse}, Sparse-Adapters~\cite{he2022sparseadapter}, etc. Among these, Sparse-Adapters~\cite{he2022sparseadapter} explores the use of popular pruning techniques (e.g., SNIP~\cite{lee2018snip}) to prune out adapters to improve their efficiency. SoRA~\cite{ding2023sparse} proposes an adaptive rank version of LoRA by gating elements of down and up projection layers and pruning out the zero entries at inference. Finally, RoSA~\cite{nikdan2024rosa} combines a sparse adapter with a low rank one to achieve some high rank benefits. However, since they combine their method with LoRA, the fused adapter weight still overwrites the entire pretrained weight tensor.

\textbf{Partial Finetuning.} Our work is most closely related to partial finetuning techniques that were mostly proposed in the pre-LoRA era~\cite{zhao2020masking, sung2021training, ansell2021composable, xu2021raise, guo2020parameter}. These methods use a mix of fixed sparse masks~\cite{sung2021training} or learned masks~\cite{zhao2020masking, guo2020parameter} to finetune a pretrained network. 
Note that, these techniques have been mostly explored for relatively \textit{small} language models, and \textit{not} for recent LLMs and diffusion models. Since the LoRA models exploded in popularity, it has been unclear if other sparse finetuning techniques would achieve comparable results to LoRA on generic adapter tasks, particularly in the vision domain. One \textit{significant} limitation of partial finetuning, as opposed to LoRA-based methods, is its \textit{high GPU memory consumption}, making it \textit{impractical} to be used for large generative models. Consequently, the reduced memory consumption for finetuning was a key factor to LoRA's success and its widespread adoption. To this end, we provide a memory- and latency-efficient PEFT-based implementation for SHiRA which trains as efficiently as LoRA, thus requiring significantly lower memory consumption compared to prior partial finetuning techniques. Further, we explore the effectiveness of sparse finetuning on both large language and vision models and provide a detailed analysis on rapid switching and multi-adapter fusion of the high rank adapters. %To our knowledge, high rank characteristics of sparse adapters has not been addressed by the partial finetuning literature.

A notable concurrent work is SpIEL~\cite{spiel} which scales partial finetuning to modern LLMs and also has a PEFT implementation that results in comparable speed and memory as LoRA. The main differences between SpIEL and SHiRA are as follows: (\textit{i})~SpIEL works with dynamic masks while SHiRA uses a static mask. (\textit{ii})~Dynamic mask in SpIEL requires users to install custom sparse linear layer kernels for the GPUs. In contrast, SHiRA does not require installing any custom kernels and directly works with native Pytorch. Hence, SHiRA's biggest advantage is its ease of training/inference deployment. (\textit{iii})~We also analyze multi-adapter fusion properties, e.g., impact of sparsity on orthogonality between adapters, which were not discussed in SpIEL. (\textit{iv})~Finally, SHiRA demonstrates its effectiveness on both vision and language tasks, whereas SpIEL only discusses the language tasks.

\textbf{Multi-Adapter Fusion.} Existing Multi-adapter fusion methods focus on preventing concept loss~\cite{gu2024mix, yu2023language, shah2023ziplora}. However, these methods usually either just use the base LoRA as it is (and then perform some non-trivial postprocessing on them)~\cite{yu2023language, shah2023ziplora}, or some create some minor variants~\cite{gu2024mix}. In contrast, we introduce a new adapter for the concept loss problem where multiple concepts naturally do not interfere with each other. In that respect, our work is orthogonal to the prior multi-adapter fusion work since our adapter can be further postprocessed using such techniques.

%% file: sections/03_method.tex
\begin{figure}[t]
  \centering
   \includegraphics[width=0.78\linewidth]{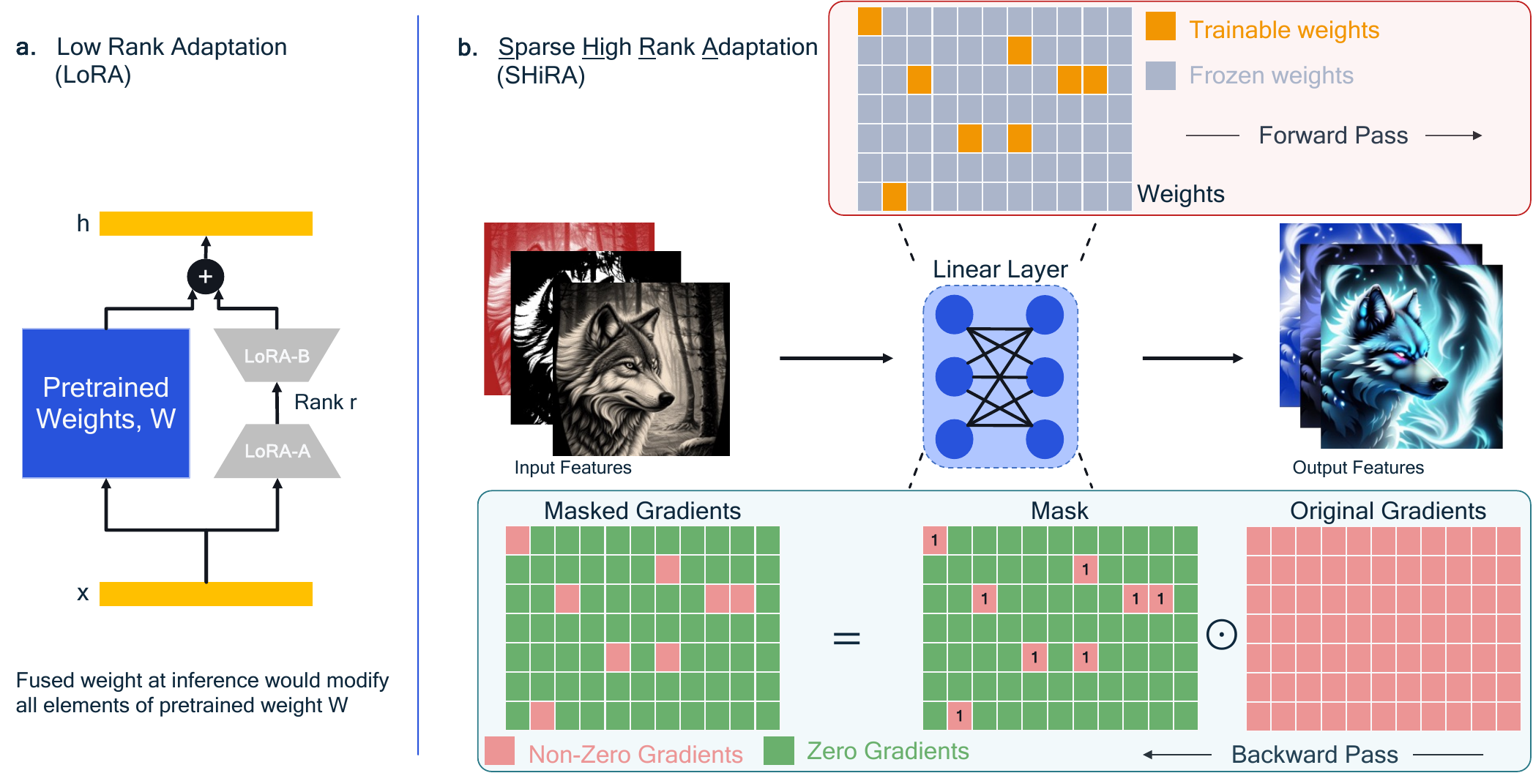}\vspace{-3mm}
   \caption{(a) LoRA when fused into the pretrained model modifies all weights and prevents rapid adapter switching. (b) SHiRA does not require additional weights during training but finetunes very few pretrained weights. Our approach relies on a sparse mask for gradient-masking during training. We show that finetuning as low as $1$-$2\%$ parameters is sufficient to achieve high accuracy.}
   \label{fig:shiraApproach}
   \vspace{-0.15cm}
\end{figure}

\vspace{-0.3cm}
\section{Proposed Approach}\label{sec::approach}
\vspace{-0.3cm}
\subsection{\underline{S}parse \underline{Hi}gh \underline{R}ank \underline{A}dapters (SHiRA)} \label{shira-approach}
\vspace{-0.2cm}
SHiRA exploits highly sparse trainable parameters in the pretrained model. In its simplest form, our adapter can be trained by masking gradients such that only a fraction of original weights get updated. Specifically, we do not add any new weights to the forward pass like LoRA (see Fig.~\ref{fig:shiraApproach}(a)) but rather make a small percentage of existing weights trainable (see Fig.~\ref{fig:shiraApproach}(b) top). To this end, we first create an extremely sparse ($\sim$$98$-$99\%$ zeros) mask $\mathcal{M}\in\mathbb{R}^{n\times m}=\{0,1\}^{n\times m}$, where $n, m$ are dimensions of the pretrained weight matrix. $\mathcal{M}$ is then used to mask the gradients during backpropagation using a Hadamard product (see Fig.~\ref{fig:shiraApproach}(b) bottom). Thus, very few parameters get updated during training and our adapter consists of just those sparse weights. Concrete gradient masking-based and another latency-/memory-efficient PEFT implementations for SHiRA are discussed in section~\ref{sec::impl}.

We consider the following masks $\mathcal{M}$ (only $1$-$2\%$ trainable parameters, see also Appendix~\ref{other-masks}):%. Note that, we only set $1$-$2\%$ of the parameters as trainable in each case:

\begin{enumerate}\setlength{\itemsep}{-0.2em}
    \item \textbf{SHiRA-Struct:} In this structured mask, certain rows or columns of the weight as well as its diagonal are set to be trainable. All other rows/columns are not trainable. The diagonal makes the mask high rank whereas the structured trainable rows/columns  -- set to 1 to enable gradient flow to corresponding parameters -- lead to a rank 1 adapter. Thus, SHiRA-Struct is a combination of a high rank but very sparse adapter and a rank 1 adapter.
    \item \textbf{SHiRA-Rand:} This mask is obtained by randomly setting $1$-$2\%$ parameters as trainable. 
    \item \textbf{SHiRA-WM:} Here we pick top-K parameters to train based on their weight magnitudes (WM), the absolute value of the weight for each layer.
    \item \textbf{SHiRA-Grad:} This is a gradient-based mask. We first collect gradients on a small calibration set and then pick top $1$-$2\%$ weights that receive the highest gradient magnitudes.
    \item \textbf{SHiRA-SNIP:} The SNIP metric from the pruning literature~\cite{lee2018snip} combines weight magnitude and gradient strategies, i.e., SNIP equals magnitude of the gradient times the weight. 
\end{enumerate} 

\subsection{Rapid Adapter Switching, Multi-Adapter Fusion, and High Rank}
\begin{figure}[t]
  \centering
   \includegraphics[width=0.93\linewidth]{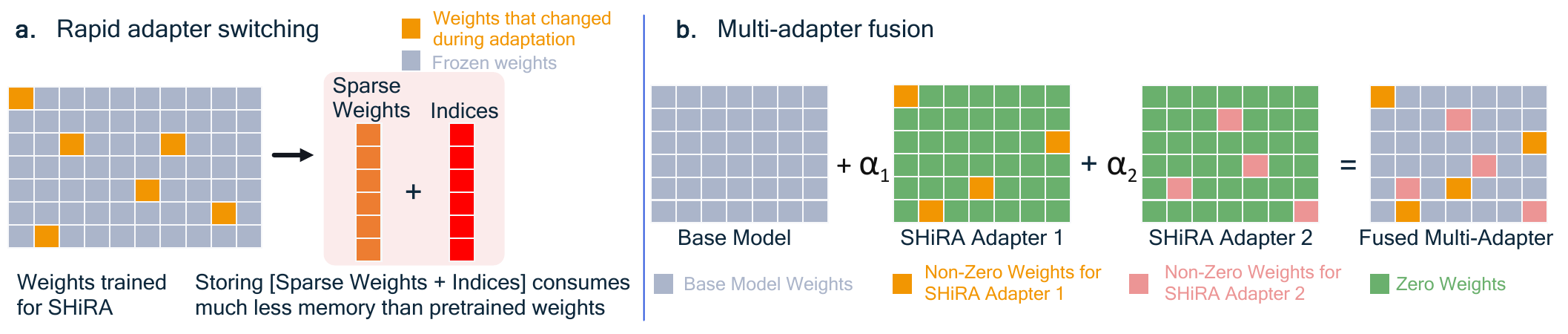}\vspace{-3mm}
   \caption{(a) Rapid adapter switching: The sparse finetuned weights can be stored as weights and their indices. At inference time, these weights can be loaded on the base model weights. Since only $1$-$2\%$ weights need to be overwritten, the adapter can be efficiently switched with different weights at inference, eliminating the need for a separate fusion stage. (b) Multi-adapter fusion: Concept-loss can be reduced if multiple adapters do not significantly interfere with each other.}
   \label{fig:shiraBenefits}
   \vspace{-0.4cm}
\end{figure}

\vspace{-0.3cm}
Since very few base weights change during the SHiRA training, we can simply extract them out and store them as sparse weights and their indices (see Fig.~\ref{fig:shiraBenefits}(a)). Hence, SHiRA is comparable to LoRA in model size but overwrites only a fraction of the pretrained weights at inference time. In contrast, LoRA fuses into base weights as $W_{new} = W + AB$ and changes the entire weight. Note that, we do not actually need to fuse SHiRA but rather just need to overwrite the modified value at the correct index in the pretrained weight tensor. This enables rapid switching on resource-constrained devices. To verify that SHiRA indeed provides rapid switching benefits compared to LoRA, we provide an optimized implementation based on \texttt{scatter\_op} to overwrite base model weights instead of fusing them like LoRA. We demonstrate that on a CPU, \textbf{weight loading for SHiRA adapters can be up to }$\bm{5\times}$-$\bm{16\times}$ \textbf{faster than equivalent LoRA fusing for inference} (see Appendix~\ref{sec::scatterop} and Fig~\ref{fig:speed}).

Next, we discuss multi-adapter fusion in SHiRA. Given two adapters $\mathcal{A}_1$ and $\mathcal{A}_2$ with sparse masks $\mathcal{M}_{1}$ and $\mathcal{M}_{2}$, we ask the following questions: (\textit{i}) What is the impact of sparsity on relative interference between adapters in the multi-adapter setting? (\textit{ii}) Is it possible to create masks that result in nearly orthogonal SHiRA weights so they do not significantly interfere with each other at inference time?

Getting adapters that do not interfere with each other is essential to avoid concept-loss. To this end, we define specific metrics in section~\ref{sec::theoryOrtho} to analyze orthogonality properties between adapter weights for various SHiRA strategies. We theoretically show that at least one of the SHiRA methods, i.e., SHiRA-Struct can in fact create near-orthogonal adapters. We further experimentally demonstrate in section~\ref{lvm-2} that SHiRA-Struct indeed outperforms other methods for multi-adapter fusion.

Finally, since we do not have any low rank weights in the forward pass, our proposed adapters can be high rank albeit highly sparse. We theoretically analyze the rank vs. sparsity properties in section~\ref{sec::theory}.

\subsection{Memory- and Latency-Efficient SHiRA Training}\label{sec::impl}\vspace{-0.3cm}
We have created two implementations for SHiRA: (\textit{i})~a backward hook-based gradient masking to turn any trainer into SHiRA finetuning (see Appendix~\ref{Latencyeff}), and (\textit{ii})~a PEFT-based implementation. As discussed in Appendix~\ref{Memoryeff}, the PEFT-based SHiRA implementation consumes \textbf{\bm{$16.63\%$} lower peak GPU memory and trains almost at a similar speed as LoRA}. On the contrary, DoRA exhibits a $40.99\%$ and $28.9\%$ increase in memory and training time respectively compared to LoRA.

%% file: sections/06_theory.tex
\vspace{-0.2cm}
\section{Theoretical Insights for SHiRA}\label{sec::theory}
\vspace{-0.2cm}
\subsection{Rank vs. Sparsity}
\vspace{-0.2cm}

\label{sec::theoryHR}
Below we discuss parameter and learning complexity, parallels between LoRA and SHiRA, as well as its optimization properties from the lens of rank and sparsity.

\begin{lemma}\label{lemmaComp}
The parameter complexity and learning complexity of SHiRA is equal to the number of non-zero elements in the adapter.\vspace{-2mm}
\end{lemma}
Appendix~\ref{prooflemmaComp} provides the proof. This lemma suggests that despite high rank property of SHiRA, it would not require significantly larger datasets to converge. 

\begin{lemma}\label{lemmaSing}
If we specify a sparsity factor, the LoRA is $r$ rank approximation of SHiRA with approximation error bounded by $\sigma_{r+1}^2$, the $(r+1)^{th}$ singular value of the SHiRA adapter.\vspace{-2mm}
\end{lemma}
The above lemma is proved in section~\ref{prooflemmaSing}. As a consequence of this lemma, any $r$ rank LoRA adapter of size $(m,n)$ can be seen as an approximation of a SHiRA adapter with $mr + rn$ non-zero elements. 

\begin{lemma}\label{lemmaScale}
Scaling factor for SHiRA is independent of the rank of the adapter and can be set to 1.\vspace{-2mm}
\end{lemma}
Please see the proof in Appendix~\ref{prooflemmaScale}. Lemma~\ref{lemmaScale} states that we do not need scaling factors to stabilize the training and, therefore, we do not need additional hyperparameters like $\alpha$ or independent learning rates for separate $A$ and $B$ matrices like in LoRA\cite{lora} or LoRA+~\cite{hayou2024lora+}. Of note, the scaling factor $\alpha$ can still be used at inference time to vary the intensity of the adapter.

\subsection{Adapter Weight Orthogonality in Multi-Adapter Fusion}\label{sec::theoryOrtho}
\vspace{-0.2cm}

In this section, we provide theoretical and empirical insights by studying properties of SHiRA and LoRA adapter designs for multi-adapter fusion.

\begin{lemma}\label{lemma-null}
Consider two adapters, $\Delta {W_{1}}$ and $\Delta {W_{2}}$. If one of the adapters, $\Delta {W_{1}}$ or $\Delta {W_{2}}$ lies in the null space of the other, then the  adapters will not interfere multiplicatively.
\vspace{-0.15cm}
\end{lemma}

Proof is given in Appendix~\ref{proof-lemma-null}. The above lemma implies that two adapters can be efficiently fused without interference if they are orthogonal. In order to analyze the orthogonality between any two adapter weights, we define the following metrics:

% \begin{Defination}
\textbf{Definition 1. \textit{Adapter Weight Orthogonality Magnitude (AWOM)}} is defined as the $l_2$ norm of the product $\mathcal{A}_1^T\mathcal{A}_2$ for two sparse adapter weights $\mathcal{A}_1, \mathcal{A}_2\in\mathbb{R}^{n\times m}$. AWOM enables us to understand how far the product $\mathcal{A}_1^T\mathcal{A}_2$ is from a zero matrix $\mathbb{O}\in \mathbb{R}^{m \times m}$ ($\mathbb{O}_{i,j}=\{0\} \forall i, j$). 

\textbf{Definition 2. \textit{Adapter Weight Orthogonality Ratio (AWOR)}} is defined as the sparsity ratio of the product $\mathcal{A}_1^T\mathcal{A}_2$. Specifically,  $\text{AWOR}=\left[1-\left(\frac{||\mathcal{A}_1^T\mathcal{A}_2||_0}{m^2}\right)\right]$, where $m^2$ is \textit{\#}elements in $\mathcal{A}_1^T\mathcal{A}_2$. 

Together, AWOM and AWOR can provide us an idea of relative orthogonality between adapter weights $\mathcal{A}_1$ and $\mathcal{A}_2$. Next, we analyze how at least one of the SHiRA strategies (i.e., SHiRA-Struct) can result in near-orthogonal adapters. Recall that, SHiRA-Struct adapters train certain rows/columns and the diagonal elements while keeping all other parameters frozen. Hence, the final trained adapter (after subtracting the pretrained weight) contains a structured pattern of rows/columns and diagonal elements, everything else being zero. Now, without loss of generality, consider two SHiRA-Struct adapters for a layer with square $m\times m$ weights: $\mathcal{A}_1 = \mathbb{I} + \mathcal{S}_1$ and $\mathcal{A}_2 = \mathbb{I} + \mathcal{S}_2$, where $\mathcal{S}_1$ and $\mathcal{S}_2$ are row-wise patterns of trained weights for two different tasks, and $\mathbb{I}$ is an identity matrix. Also, $\mathcal{S}_1$ and $\mathcal{S}_2$ are non-overlapping, e.g., both have same number of non-zero rows but are offset from each other such that they do not have any common trained rows. Then, the following result holds:

\begin{lemma}\label{lemma-struct}
    Non-overlapping SHiRA-Struct adapters are nearly orthogonal: AWOR for non-overlapping SHiRA-Struct adapters is at most the sum of sparsity of individual adapters. Since all SHiRA masks are highly sparse, $\mathcal{A}_1^T\mathcal{A}_2$ has a lot of zeros, thus making the adapters nearly orthogonal.
\end{lemma}
\vspace{-0.15cm}
Proof is provided in Appendix~\ref{proof-lemma-struct}.

\begin{wrapfigure}{r}{0.55\textwidth}
    \begin{center}\vspace{-4mm}
               \includegraphics[width=\linewidth]{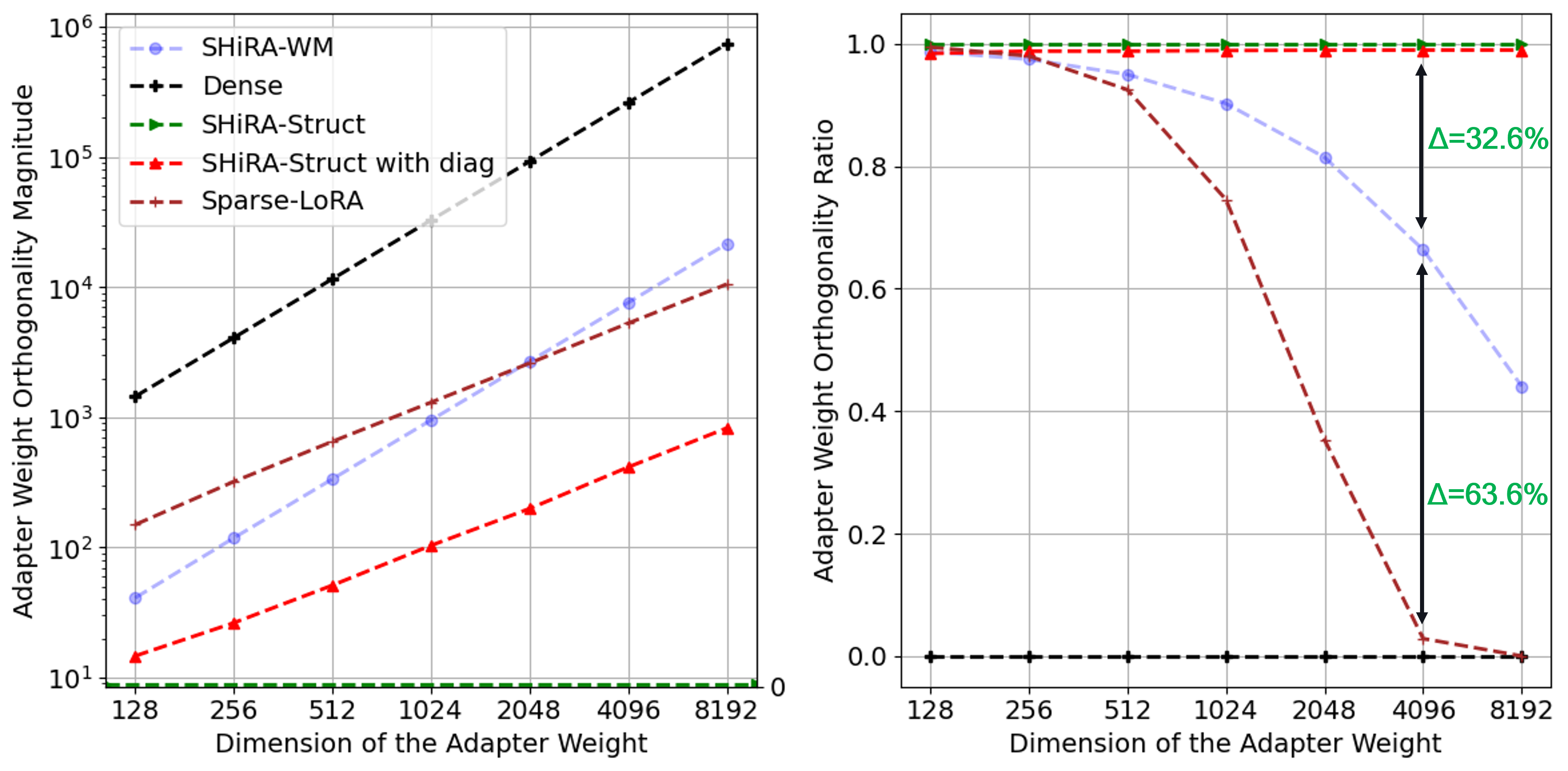}\vspace{-7mm}
       \caption{Comparison of average AWOM \textit{(left)} and AWOR \textit{(right)} for 50 randomly initialized adapters. We compare different adapters, namely - Dense, Sparse LoRA, SHiRA-WM and SHiRA-Struct.}
       \label{fig:AWOR-AWOM}
    \end{center}
    \vspace{-6mm}
\end{wrapfigure}

We demonstrate the orthogonality properties of various adapters and report the simulation results in Fig.~\ref{fig:AWOR-AWOM}. For our experiment, we compute AWOM and AWOR for a variety of adapter designs - 
dense, sparse-LoRA~\cite{he2022sparseadapter} (sparse LoRA A and B weights), SHiRA-WM and SHiRA-Struct based adapters. As shown in Fig.~\ref{fig:AWOR-AWOM}, both dense and sparse LoRA have low AWOR for adapters with larger dimensions, e.g., 4096 $\times$ 4096 which is typical in LLMs. This signifies that these adapter weights are non-orthogonal. On the contrary, SHiRA-WM achieves much higher AWOR than the LoRA variants. More interestingly, SHiRA-Struct is nearly orthogonal. Note that, due to high sparsity, AWOM also tends to be much lower for SHiRA adapters than the dense counterparts. Combined with the fact that AWOR of SHiRA adapters is 63-96\% higher sparsity than LoRA, this may suggest that  $\mathcal{A}_1^T\mathcal{A}_2$ would be closer to zero for SHiRA adapters, thus potentially bringing them closer to orthogonality and less interference. Finally, although we have shown interesting properties for SHiRA-Struct, it is still a rank 1 + diagonal adapter. Hence, we need to tradeoff single adapter performance (which strongly depends on adapter's expressive power) against the multi-adapter fusion capabilities. For instance, next we will see that while SHiRA-Struct is good for vision, SHiRA-SNIP performs well across both LVMs and LLMs.

\textbf{Remark 1.} The orthogonality property shown here can lead to disentangled representation for adapter outputs before they merge into the base model. However, this property does not hold for other SHiRA masks that do not have a regular sparsity pattern like SHiRA-Struct even if other SHiRA strategies are still more orthogonal than LoRA weights (e.g., see SHiRA-WM AWOR in Fig.~\ref{fig:AWOR-AWOM}(right)). Interestingly, for unstructured sparse masks like SHiRA-WM, SHiRA-Grad, SHiRA-SNIP, etc., both overlapping and non-overlapping adapters have similar orthogonality properties. We discuss this in more detail in section~\ref{mafLLMres}. Finally, this analysis only focuses on \textit{orthogonality of adapter weights} and \textit{not} on orthogonality of subspaces. We leave the subspace analysis of SHiRA for future work.

%% file: sections/04_experiments.tex
\vspace{-0.4cm}
\section{Experiments}\label{sec::exp}
\vspace{-0.3cm}
%In this section, we showcase the effectiveness of SHiRA and conduct extensive evaluation on various tasks and benchmarks in both vision and language domains. 

\subsection{Training Setup and Datasets}
\vspace{-0.3cm}
For the vision tasks, we use the RealisticVision-v3 model checkpoint for Stable Diffusion-v1.5, and finetune it using different adapters on two style transfer datasets collected using public domain images. The first dataset is called Bluefire which provides a ``blue fire'' effect to images. The second dataset is a painting dataset which gives a ``paintings'' effect (see Appendix section \ref{dataset-desp} for more details). For both these datasets, we conduct single- and multi-adapter experiments. To quantify the image quality, we use the Human Preference Score-V2 (HPSv2)~\cite{wu2023human}.

On the language domain, we experiment with LLaMA 7B~\cite{llama}, LLaMA2-7B~\cite{llama2} and evaluate it on various commonsense reasoning benchmarks such as HellaSwag, PIQA, SIQA, BoolQ, Arc-easy, Arc-challenge, OpenBookQA and Winogrande. Similar to our vision investigations, we conduct single- and multi-adapter experiments on LLMs as well. Specifically, for language finetuning, we follow the setup adopted by \cite{llmadapters, dora} for training and evaluating LoRA~\cite{lora}, DoRA~\cite{dora}, and SHiRA based finetuned models on downstream tasks.

Finally, we also explore generalizability of SHiRA to other popular LoRA models and applications such as SDXL~\cite{sdxl} and DreamBooth~\cite{ruiz2023dreambooth}. Detailed training setups are provided in the Appendix~\ref{training-details}.

\vspace{-3mm}
\subsection{Vision Results}
\vspace{-0.2cm}
\subsubsection{Impact of Various SHiRA Masks}\label{single-shira-lvm}
\vspace{-0.2cm}
We first evaluate the image quality for SHiRA and LoRA on Paintings and Bluefire datasets for both single and multi-adapter usecases. Fig.~\ref{fig:introFigure} demonstrates comparison between SHiRA-SNIP and LoRA. As evident, by merely changing 2\% pretrained weights, SHiRA generates high quality images for both finetuning tasks. 
\begin{wraptable}{r}{8.3cm}
    %\addtolength{\tabcolsep}{-2pt}
    \centering\vspace{-1.5mm}
    \fontsize{7.0pt}{5.75pt}\selectfont
    % \fontsize{8.00pt}{8.25pt}\selectfont
\scalebox{0.95}{
    \begin{tabular} {  lcccccccc }
    \toprule
    
    \multirow{1}{*}{\textbf{Style}} & \multirow{1}{*}{\textbf{Method}}  & \textbf{\%Params} & \multicolumn{2}{c}{\textbf{HPSv2 score($\uparrow$)}} \\

    & & & $\alpha=1$ & $\alpha=0.5$ \\
    
    \hline
    \midrule
    \multirow{6}{*}{\vspace{0.1in}Paintings} & LoRA & 3.84 &$24.7 \pm1.8$ & $31.3 \pm 1.5$ \\
    & SHiRA-Struct &  \textbf{1.99} &$\bm{31.2 \pm 1.7}$ &  $\bm{33.0 \pm 1.8}$ \\
    % & SHiRA-Rand & 2.05&$30.7 \pm 1.9$ &  $32.7 \pm 1.9$ \\
    % & SHiRA-WM & 2.05&$29.7 \pm 1.9$ & $32.1 \pm 1.8$ \\
    & SHiRA-Grad & 2.05&$30.3 \pm 1.8$  & $32.3 \pm 1.8$ \\
    & SHiRA-SNIP & 2.05&$29.8 \pm 1.8$ & $31.6 \pm 1.8$ \\
    \midrule
    \multirow{6}{*}{\vspace{0.1in}Bluefire} & LoRA & 3.84&$32.6 \pm 1.9$ &  $33.6 \pm 1.6$ \\
    & SHiRA-Struct & \textbf{1.99} &$\bm{34.2 \pm 1.6}$  & $\bm{34.1 \pm 1.5}$  \\
    % & SHiRA-Rand & 2.05&$33.4 \pm 1.9$  & $33.7 \pm 1.7$ \\
    % & SHiRA-WM & 2.05&$31.9 \pm 2.1$  & $33.1 \pm 1.7$ \\
    & SHiRA-Grad & 2.05&$\bm{34.2 \pm 1.5}$ & $33.7 \pm 1.7$ \\
    & SHiRA-SNIP & 2.05&$33.7 \pm 1.7$  & $33.7 \pm 1.6$ \\
    \bottomrule

\end{tabular}\vspace{-3mm}
}
    \caption{HPSv2 score of various adapters on Paintings and Bluefire. SHiRA-Struct outperforms all other methods.} 
    \label{tab:mainVisiontable}
\vspace{-0.5cm}
\end{wraptable}
Next, we compare various types of SHiRA masks in Fig.~\ref{fig:shiraFigure}. Clearly, all SHiRA schemes produce impressive images for different prompts and significantly outperform LoRA. We further quantify the image quality using HPSv2 for each of the masks. The results are presented in Table~\ref{tab:mainVisiontable}. As evident, all variants of SHiRA consistently achieve superior or similar HPSv2 scores than LoRA, especially for larger $\alpha$ (see details on scaling factor $\alpha$ in Appendix~\ref{alpha-effect}). More results are provided in Appendices~\ref{tab11} and~\ref{AppMoreRes}: see Table~\ref{tab: t2i_table} and Fig. ~\ref{fig:appendixfig}, ~\ref{fig:shiraFigure2},~\ref{fig:shiraFigure3}.

\vspace{-3mm}
\subsubsection{SHiRA Adapters aid Multi-Adapter Fusion}\label{lvm-2}
\vspace{-0.2cm}
As explained in section~\ref{sec::theoryOrtho}, high sparsity of SHiRA reduces their AWOM and increases the AWOR metrics by increasing the number of zeros in $\mathcal{A}_1^T\mathcal{A}_2$ product even for unstructured schemes such as SHiRA-WM, SHiRA-Grad, and SHiRA-SNIP. We hypothesized that this may lead to improved multi-adapter fusion performance. This was also pointed out by~\cite{shah2023ziplora,gu2024mix,wang2023orthogonal}: naively merging multiple LoRA adapters leads to poor performance and concept loss. 

\begin{figure}[t]
  \centering
   \includegraphics[width=\linewidth]{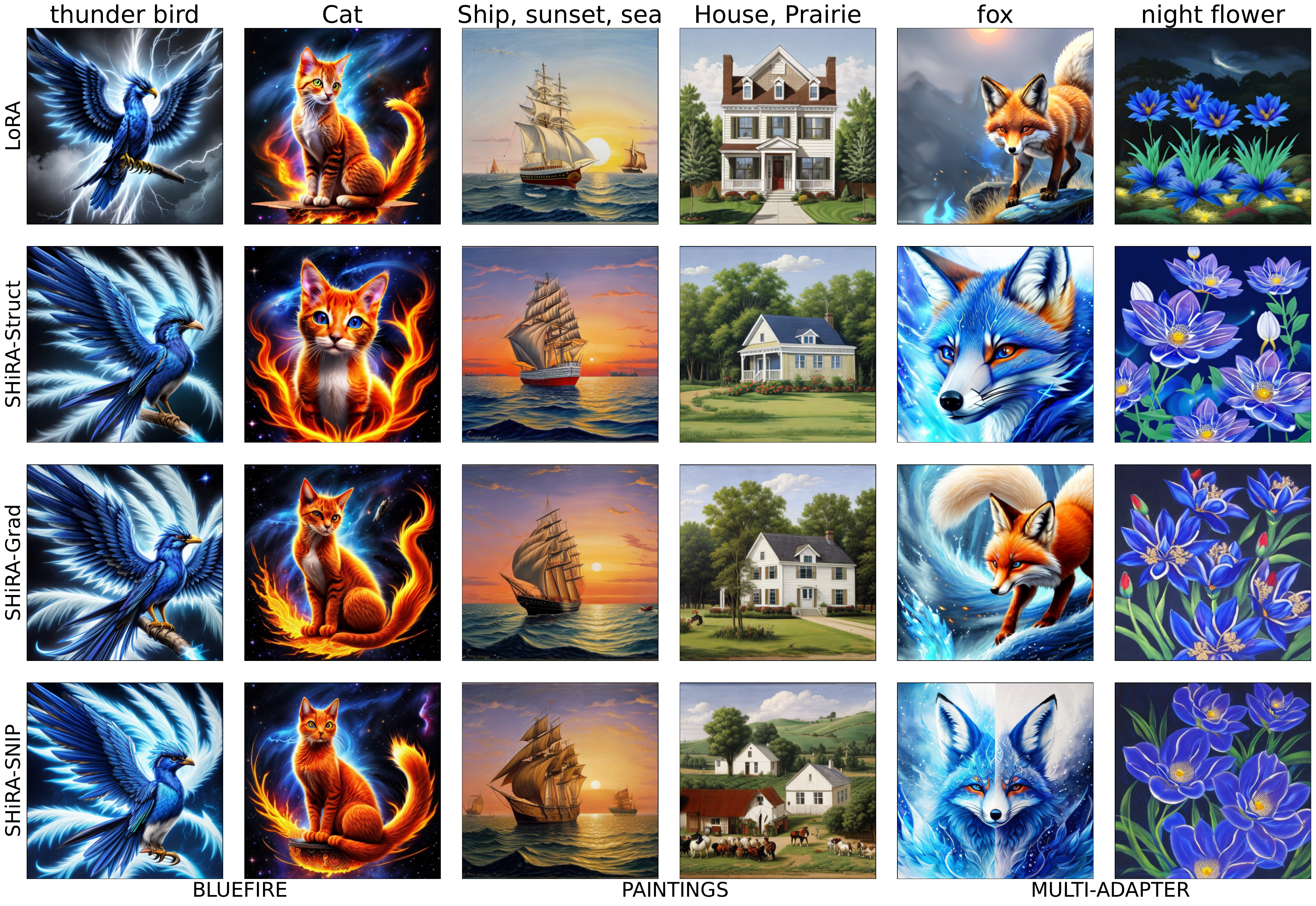}\vspace{-6mm}
   \caption{Comparison between different SHiRA masking methods for single- and multi-adapter image generation. For multi-adapter fusion, SHiRA-Struct outperforms all other adapters by generating exceptional images with high frequency details and good concept fusion (e.g., see fox and flower).}
   \label{fig:shiraFigure}\vspace{- 0.5 em}
\end{figure}

We now validate the effectiveness of various SHiRA schemes on multi-adapter fusion. The right two columns in Fig.~\ref{fig:introFigure} and Fig.~\ref{fig:shiraFigure} show our results. SHiRA is clearly better at capturing both concepts than LoRA. For example, both bird and knight images in Fig.~\ref{fig:introFigure} generated with LoRA lose most of the paintings concept. Similarly, for the fox image in Fig.~\ref{fig:shiraFigure}, LoRA does not show significant bluefire concept. In contrast, SHiRA-Struct and SHiRA-SNIP consistently perform well on many different prompts and produce exceptional images for multi-adapter fusion. Please refer to Appendix~\ref{AppMoreSamplesLVM} (Fig.~\ref{fig:appendixfig},~\ref{fig:shiraFigure2},~\ref{fig:shiraFigure3}, and ~\ref{fig:shiraFigure4}) for additional results. For certain classes that were not included in the training set for both adapters (e.g., see Koala in Fig.~\ref{fig:appendixfig},~\ref{fig:shiraFigure3}, and~\ref{fig:shiraFigure4} in Appendix), we observe that LoRA produces significant artifacts whereas SHiRA generates high quality images.

\vspace{-3mm}
\subsection{Language Results}
\vspace{-0.2cm}

\subsubsection{Single Adapter SHiRA Finetuning}\label{single-adapter-llm}
\vspace{-0.2cm}
Similar to vision results, we demonstrate the effectiveness of SHiRA on language tasks. For our experiments, each adapter (i.e., weight-magnitude, gradient-magnitude, and SNIP based SHiRA) is trained on the combined 170K sample commonsense reasoning dataset released by~\cite{llmadapters, dora}. Similar to~\cite{dora}, we train our SHiRA adapters for 3 epochs and compare it against the LoRA baselines. As shown in Table~\ref{tab: CommonSense-Single-Adapter}, various SHiRA adapters outperform LoRA by 1.9-2.7\% on an average on LLaMA-7B. Importantly, SHiRA only modifies 1\% base parameter weights as compared to \textbf{66.72\%} (\textbf{4.5B weights}) changed by LoRA in the fused mode, thus enabling rapid switching on edge devices. Interestingly, we found that SHiRA-Struct does not perform well on language tasks likely because it is a rank 1 + diagonal adapter and may not have sufficient expressive power.

Moreover, when compared to newer techniques like DoRA~\cite{dora}, our proposed work takes an orthogonal approach by finetuning very few parameters of the pretrained weights. This strategy allows for an efficient integration of our adapter with methods like DoRA to improve the expressiveness of the adapters. As we show in Table~\ref{tab: CommonSense-Single-Adapter}, our proposed adapter benefits from DoRA based finetuning and achieves almost comparable performance (within 0.3\%) to DoRA on an average, with an added benefit of changing only 1\% parameters at inference time. In contrast, DoRA would lead to \textbf{66.72\%} (\textbf{4.5B weights} $\approx$ \textbf{9GB memory} in FP16 format) parameter change in the fused mode. Therefore, SHiRA is orthogonal to other existing low rank methods and can be efficiently integrated with them.\\
Finally, we experiment with LLaMA2-7B~\cite{llama2} and demonstrate that SHiRA-SNIP -- which achieved the best results on LLaMA-7B -- yields significant accuracy gains compared to LoRA and nearly the same accuracy as DoRA (within 0.4\%, see Table~\ref{tab: CommonSense-2}). 

\begin{table*}[t]        
    \scalebox{0.9}{
    \addtolength{\tabcolsep}{-2pt}
    \fontsize{7.0pt}{5.75pt}\selectfont
    % \fontsize{8.00pt}{8.25pt}\selectfont
    \begin{tabular} {  lccccccccccc }
    \toprule
    \textbf{Model} & \textbf{\%Params} & \textbf{\%C} & \textbf{BoolQ($\uparrow$)} & \textbf{PIQA($\uparrow$)} & \textbf{Arc-e($\uparrow$)} & \textbf{Arc-c($\uparrow$)} & \textbf{WG($\uparrow$)} & \textbf{OBQA($\uparrow$)} & \textbf{HS($\uparrow$)} &  \textbf{SIQA($\uparrow$)} &\textbf{Avg.($\uparrow$)} \\\hline
    \midrule
    LoRA & \textbf{0.83} & {\color{red} 66.72} & 68.9 & 80.7 & 77.8 & 61.3 & 78.8 & 74.8 & 78.1 & 77.4 & 74.7 ({\color{red} +0\%}) \\
    \midrule
    SHiRA-Grad  & 1.0 & {\color{ForestGreen}\textbf{1.0}} & 68.4 & 80.9 & 80.2 & 64.7 & \textbf{80.4} & 78.2 & 80.3 & \textbf{79.4} & 76.6 ({\color{ForestGreen} +1.9\%})\\
    \midrule
     SHiRA-WM  & 1.0 & {\color{ForestGreen}\textbf{1.0}} & \textbf{69.6} & \textbf{81.6} & \textbf{81.5} & 66.5 & 79.8 & 79.4 & 79.6 & 77.8 & 77.0 ({\color{ForestGreen} +2.3\%})\\
    \midrule
    \textbf{SHiRA-SNIP}  & 1.0 & {\color{ForestGreen}\textbf{1.0}} & 68.3 & 80.6 & \textbf{81.5} & \textbf{67.9} & 80.0 & \textbf{79.6} & \textbf{82.1} & 79.1 & \textbf{77.4} ({\color{ForestGreen} +2.7\%})\\
     \hline %\\ 
    \midrule
    %\bottomrule
    DoRA & 0.84 & {\color{red} 66.72} & 68.5 & \textbf{82.9} & 81.4 & \textbf{65.8} & \textbf{80.8} & \textbf{81.0} & \textbf{84.8} & \textbf{79.6} & \textbf{78.1} ({\color{ForestGreen} +0\%})\\
    \midrule
    SHiRA-WM-DoRA & 6.25$^*$ & {\color{ForestGreen}\textbf{1.0}} & \textbf{70.9} & 81.9 & \textbf{81.7} & 64.9 & \textbf{80.8} & 79.2 & 84.5 & 78.6 & \textbf{77.8} ({\color{red} -0.3\%})\\
    \bottomrule
\end{tabular}\vspace{-5mm}
}
    \caption{Evaluation of LLaMA-7B on Commonsense Reasoning. WG and HS denote WinoGrande and HellaSwag, respectively. \%C represents parameters changed in the fused mode.  ($\uparrow$): the higher the better. {\color{ForestGreen} Green} denotes improvement. $^*$Trained by masking a high-rank DoRA with a WM mask of top 1\% weights, thus changing only 1\% of the model during both training and inference.} 
    \label{tab: CommonSense-Single-Adapter}
% \vspace{-0.5cm}
\end{table*}

\begin{table*}[t]
\centering
\scalebox{0.925}{
    \addtolength{\tabcolsep}{-2pt}
    \fontsize{7.0pt}{5.75pt}\selectfont
    % \fontsize{8.00pt}{8.25pt}\selectfont
    \begin{tabular} {  lccccccccccc }
    \toprule
    \textbf{Model} & \textbf{\%Params} & \textbf{\%C} & \textbf{BoolQ($\uparrow$)} & \textbf{PIQA($\uparrow$)} & \textbf{Arc-e($\uparrow$)} & \textbf{Arc-c($\uparrow$)} & \textbf{WG($\uparrow$)} & \textbf{OBQA($\uparrow$)} & \textbf{HS($\uparrow$)} &  \textbf{SIQA($\uparrow$)} &\textbf{Avg.($\uparrow$)} \\\hline
    \midrule
    LoRA & \textbf{0.83}& {\color{red}66.72} & 69.90 & 79.9 & 79.8 & 64.7 & 82.6 & 81.0 & 83.6 & \textbf{79.5} & 77.61 ({\color{red} +0\%})\\
    \midrule
    DoRA & 0.84 & {\color{red}66.72} & \textbf{71.8} & \textbf{83.7} & \textbf{83.7} & 68.2 & \textbf{82.6} & \textbf{82.4} & 89.1 & 76.0 & \textbf{79.68} ({\color{ForestGreen} +2.07\%})\\
    \midrule
    \textbf{SHiRA-SNIP}  & 1.0 & {\color{ForestGreen}\textbf{1.0}} & 70.42 & 81.71 & 83.25 &  \textbf{68.6} & 80.51 & 81.0 &  \textbf{89.78} & 79.01 & \textbf{79.28} ({\color{ForestGreen} +1.67\%}) \\
    
    \midrule
\end{tabular}\vspace{-5mm}
}
    \caption{Results for LLaMA2-7B on Commonsense Reasoning.}
    \label{tab: CommonSense-2}
\end{table*}

\vspace{-0.3cm}
\subsubsection{Multi-Adapter Fusion on LLMs}\label{mafLLMres}
\vspace{-0.2cm}
We now extend our LLM experiments to the multi-adapter fusion setting. To this end, we create a \textit{new} setup where we independently train multiple adapters on training sets of individual commonsense reasoning benchmarks, i.e., one adapter each for BoolQ, PIQA, and Arc-Easy. In contrast, each adapter in section~\ref{single-adapter-llm} was trained on a combined dataset containing 170K samples from all eight commonsense benchmarks as proposed in~\cite{llmadapters, dora}. In the present section, the goal is to evaluate how much accuracy drop various adapters experience when we perform multi-adapter fusion. Due to its simplicity towards constructing a mask, we will use SHiRA-WM in the rest of this paper. Further, we explore two settings - overlapping and non-overlapping SHiRA-WM adapters. The overlapping mask consists of top 1\% parameters being trained for all tasks. On the other hand, the non-overlapping setting trains the top 1\% weights for the first task, next top 1\% for the second task, and so on. We compare the performance of both LoRA and SHiRA across the multi-adapter fusion of these three tasks.
\begin{table*}[t]
    \addtolength{\tabcolsep}{-2pt}
    \centering
\scalebox{0.93}{
    \fontsize{7.0pt}{5.75pt}\selectfont
    % \fontsize{8.00pt}{8.25pt}\selectfont
    \begin{tabular} {  lcccccccccc }
    \toprule
    \multicolumn{1}{c}{} & \multicolumn{4}{c}{\textbf{Single Adapter}} & & \multicolumn{4}{c}{\textbf{Multi-Adapter}} &\\
    \cmidrule{2-5} \cmidrule{7-10}
    %\midrule
    \textbf{Model} & \textbf{BoolQ($\uparrow$)} & \textbf{PIQA($\uparrow$)} & \textbf{Arc\_e($\uparrow$)} & \textbf{Avg($\uparrow$)}&  & \textbf{BoolQ($\uparrow$)} & \textbf{PIQA($\uparrow$)} & \textbf{Arc\_e($\uparrow$)}& \textbf{Avg($\uparrow$)} & \textbf{\%Drop ($\downarrow$)}\\\hline
    \midrule
    LoRA & \textbf{80.52} & 79.05 & 75.67 & 78.41 & & 77.22 & 71.27 & 57.45 & 67.33 ({\color{red} +0\%}) & {\color{red} 11.08} \\
    \midrule
    SHiRA-WM-Overlap & 78.07 & \textbf{79.71} & \textbf{77.57} & \textbf{78.45}& & \textbf{77.43} & 76.88 & 67.76 & \textbf{74.02} ({\color{ForestGreen} +6.69\%}) & {\color{ForestGreen} 4.43}\\
    \midrule
    SHiRA-WM-Non-Overlap & 76.94 & \textbf{79.71}  & 75.97 & 77.54 & & 74.22  & \textbf{78.4} & \textbf{69.15} & 73.92 ({\color{ForestGreen} +6.59\%}) & {\color{ForestGreen} \textbf{3.62}}\\
    \bottomrule
\end{tabular}\vspace{-5mm}
}
    \caption{Multi-adapter fusion evaluation of independently trained SHiRA and LoRA adapters on  BoolQ, PIQA, and Arc-Easy. \%Drop is calculated as drop in average accuracy for multi-adapter fusion compared to the single adapter average accuracy for each adapter.} 
    \label{tab: multishira}
\vspace{-0.1cm}
\end{table*}
As shown in Table~\ref{tab: multishira}, both overlapping and non-overlapping multi-SHiRA outperform multi-LoRA on all three commonsense benchmarks. This is inline with our theoretical analysis in section~\ref{sec::theoryOrtho} where we suggest that even unstructured sparse SHiRA adapters such as SHiRA-WM would have more orthogonal behavior than LoRA due to high sparsity (see higher AWOR of SHiRA-WM in Fig.~\ref{fig:AWOR-AWOM}(right)). In comparison, independently trained LoRA adapters would have no such property and suffer greatly during multi-adapter fusion. As a result, we see that both SHiRA models outperform LoRA by more than 6.5\% accuracy on average. Further analysis of the properties of these trained adapters is discussed in Appendix~\ref{sec:analysis-adapters} (see Table~\ref{tab:L2} and Fig.~\ref{orthogonality-figure}).

Of note, this experiment also demonstrates the value of creating a good mask for single adapter performance: Non-overlapping masks achieve lower single adapter accuracy than the corresponding overlapping masks since they train less important parameters. Hence, creating an optimal mask for SHiRA should be of significant interest to future research. 

\vspace{-3mm}
\subsection{Content/Style Personalization: Generalizing SHiRA to SDXL and DreamBooth}
\vspace{-0.2cm}
Finally, we extend SHiRA to focus on DreamBooth~\cite{ruiz2023dreambooth} using a much bigger vision model called SDXL~\cite{sdxl}. 
\begin{figure}[t]
  \centering
   \includegraphics[width=1\linewidth]{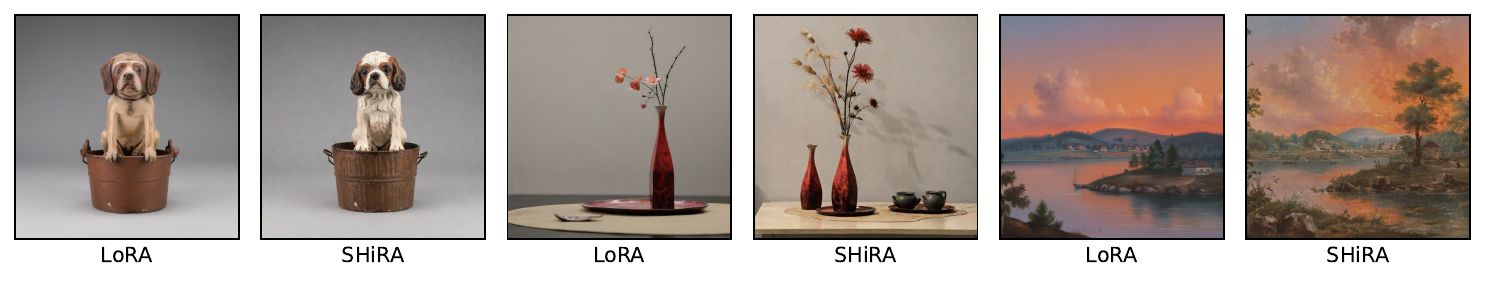}\vspace{-8mm}
   \caption{LoRA- vs. SHiRA-based DreamBooth on SDXL. Prompts for content/style personalization - \textit{left pair}: "A picture of a dog in \texttt{<STYLE:WOODEN-SCULPTURE>} style in a bucket", \textit{center pair}: "A picture of a \texttt{<CONTENT:VASE>}  with flowers", and \textit{right pair}: "A picture of a sunset in \texttt{<STYLE:CANVAS>} style". Here, "\texttt{<CONTENT>}" and "\texttt{<STYLE>}" are special identifier tokens for content/style.}
   \label{fig:dreambooth-shira}
   \vspace{-3mm}
\end{figure}
We follow a similar setup as adopted by~\cite{dreambooth-lora}. Specifically, one content (vase) and two style (wooden sculpture and canvas) datasets with five images each were collected from the DreamBooth dataset~\cite{ruiz2023dreambooth} and public domains, respectively. These datasets were used to train various content and style adapters. For our experiments, we use SDXL~\cite{podell2023sdxl} as our base model and train both LoRA and SHiRA adapters with comparable trainable parameters on individual single-concept datasets. During training, prompts containing special identifier tokens like "\texttt{<CONTENT>}" or "\texttt{<STYLE>}" (e.g., \texttt{<SBU>} as content token for vase and \texttt{<SZN>} as style token for wooden sculpture and canvas) are used to finetune the SDXL network for content or style personalization, respectively. During inference, similar prompts are used to generate images from LoRA- or SHiRA-based DreamBooth.

Fig~\ref{fig:dreambooth-shira} shows DreamBooth generated images for LoRA and SHiRA. Clearly, our proposed adapter produces high quality personalized images of target concept in different scenarios. This highlights the broad applicability of our adapter while still preserving the benefits of rapid adapter switching.\vspace{-2mm}

%% file: sections/05_conclusion.tex
\vspace{-2mm}
\section{Discussion}\label{sec::discussion}
\vspace{-3mm}
To summarize our main contributions, we highlight that SHiRA -- when used with even the most basic pruning metrics (such as weight- or gradient-magnitude, SNIP, structured masks, etc.) -- significantly outperforms LoRA on a variety of large-scale tasks in both large vision and large language domains. For LVM style transfer applications, we found that SHiRA-Struct is the most effective masking technique due to its special orthogonality properties that aid multi-adapter fusion. However, SHiRA-SNIP and SHiRA-Grad are not too far behind and achieve competitive performance as SHiRA-Struct. On the LLM commonsense reasoning side, SHiRA-SNIP is the best strategy out of the masking techniques we have considered in this work. Specifically, SHiRA-Struct did not achieve good results on the more complex commonsense reasoning tasks since it is a combination of a rank-1 + a highly sparse diagonal adapter. SHiRA-Grad on LLMs is about 0.8\% worse accuracy than SHiRA-SNIP (76.6\% vs. 77.4\% average accuracy on commonsense reasoning for LLaMA-1). Therefore, in conclusion, for the applications/fields and the masking techniques considered in this paper, SHiRA-SNIP works well across both language and vision domains. Hence, we recommend that SHiRA-SNIP is one of the strongest candidates that we have considered for sparse finetuning.

\vspace{-2mm}
\section{Conclusion}\label{sec::conclusion}
\vspace{-3mm}
In this paper, we have proposed SHiRA, a new high rank adapter paradigm to demonstrate that even finetuning merely 1-2\% parameters of the pretrained generative models is sufficient to achieve high performance on many adapter tasks. We have demonstrated SHiRA's ability to rapidly switch adapters and to avoid concept loss with support from both theory and experiments. Furthermore, we have shown how specially designed sparse masks can lead to near-orthogonal adapter weights which allows for natural multi-adapter fusion. We have conducted extensive single- and multi-adapter experiments on several vision and language tasks to demonstrate the superiority of SHiRA over LoRA. Our latency- and memory-efficient PEFT-based implementation for training SHiRA runs at nearly the same speed as LoRA while consuming about 16\% lower peak GPU memory. Finally, for inference, we have provided a \texttt{scatter\_op} based method that can load our SHiRA $5\times$-$16\times$ faster than equivalent LoRA fusion on a CPU, thus demonstrating our rapid switching benefits.

\vspace{-2mm}
\section*{Acknowledgments}\label{sec::ack}
\vspace{-3mm}
We thank anonymous reviewers for insightful comments and constructive feedback which significantly improved the quality of our work.

%% file: sections/07_appendix.tex
\section{Edge Deployment Challenges for LoRA (Contd.)}\label{sec::approachMobileContd}
To understand the overhead of each of the stages to the standard huggingface LoRA inference pipeline (i.e., \texttt{load}, \texttt{fuse}, \texttt{unfuse}, \texttt{unload}), we experiment with the pipeline 
\begin{wraptable}{r}{5.3cm}
    \addtolength{\tabcolsep}{-2pt}
    \centering
    \fontsize{6.75pt}{6.75pt}\selectfont
    
    \begin{tabular} {  lccc }
    \toprule
    \centering \textbf{Stage} & \textbf{Server-GPU (s)} & \textbf{Desktop-CPU (s)} \\\hline
    \midrule
    \texttt{load} &  {\color{red} $0.883\pm 0.085$} & $0.786\pm 0.056$\\
    \texttt{fuse} & $0.306\pm 0.044$ & {\color{red} $3.003\pm 0.023$}\\
    \texttt{unfuse} & $0.206\pm 0.041$ & {\color{red} $2.916\pm 0.014$}\\
    \texttt{unload} & $0.007\pm 0.001$ & $0.007\pm 0.001$\\
    \bottomrule
\end{tabular}\vspace{-1mm}

    \caption{Latency (in s) to \texttt{load}, \texttt{fuse}, \texttt{unfuse}, \texttt{unload}~\cite{hflora} adapters on SDXL on Server-GPU and Desktop-CPU. On a mobile device, fusing/unfusing would happen for each layer iteratively since we cannot store all weights at the same time on local on-chip memory (unlike a large GPU), resulting in much higher overhead.}
    \label{tab:hflora}
\end{wraptable}
provided in~\cite{hflora} and iteratively add adapters to SDXL model~\cite{sdxl}. As evident from Table~\ref{tab:hflora}, on a server-grade GPU, \texttt{load} time dominates whereas \texttt{fuse}/\texttt{unfuse}/\texttt{unload} times are relatively negligible. However, if we try to run the exact same pipeline on an everyday device like a desktop-grade CPU, we see that the \texttt{fuse} and \texttt{unfuse} times start dominating and can hinder rapid adapter switching. Note that, on an even more constrained device like a mobile phone, AI accelerators do not have sufficient memory to store weights from all layers at the same time in the local memory. Hence, on such devices, we would need to load base model weights for each layer into the local memory, and then fuse corresponding LoRA weights before we can run inference for that layer. This obviously leads to a massive inference latency overhead. As a result, existing deployment options are \textit{not} feasible for rapid switching on mobile devices.

\section{More Details on SHiRA Masks} \label{other-masks}
Selecting important salient weights pertinent to a task can be done in many ways, and one popular approach is to use masks. In this section we discuss various strategies to construct sparse mask based on different heuristics to select weights for efficient finetuning of large generative models.  

\subsection{Structured Sparse Mask (SHiRA-Struct)} 
This is a simple structured mask. We begin with making every $f$ rows or columns in a weight matrix trainable, where we call $f$ as the \textit{frequency} parameter and we choose it based on how much sparsity we need in the adapter. That is, the mask $\mathcal{M}$ consists of every $f$ rows or columns containing ones and everything else as zeros. This actually makes it a rank $1$ mask because all rows and columns would be linearly dependent. Therefore, to make it high rank, we also add a \textit{diagonal} parameter which makes the resulting mask $\mathcal{M}$ high rank.

\subsection{Unstructured Sparse Random Mask (SHiRA-Random)} 
Unstructured sparse random masks involve masking individual weights without any specific pattern or structure. The masked weights are randomly scattered throughout the weight tensor, resulting in a sparse weight tensor. However, as the weights are selected without considering their salience to the task, randomly selected unstructured masks may often be sub-optimal for finetuning. One common way of constructing random sparse marks is using Bernoulli sampling:
\begin{equation}
  f(k;p) =
  \begin{cases}
    p     & \text{if $k = 1$}, \\
    1 - p & \text{if $k = 0$}.
  \end{cases}
\end{equation}

where, $p$ is the probability of sampling 1 from the distribution.

\subsection{Weight Magnitude-Based Sparse Mask (SHiRA-WM)} 
Many earlier works \cite{lee2020layer,sun2023simple} have shown the importance of weight magnitude based masks for identifying important weights in the network. Motivated by this literature, we design a weight magnitude based proxy to adapt the behavior of the pretrained network. Specifically, we create a mask by choosing the top-$K$ weight magnitudes at specific layers where SHiRA is employed. We finetune only these top-$K$ weights and keep the rest of them frozen to their pretrained values. Typically, $K$ is a very small percentage of parameters so that the overall number of parameters to be tuned stays comparable to LoRA and its variants. 

\subsection{Gradient Based Sparse Mask (SHiRA-Grad)} 
Despite the efficacy of employing weight magnitude based scheme, this approach lacks an inherent awareness of the specific task for which the model is being finetuned. To address this challenge, we design a similar gradient magnitude based proxy to identity important top-$K$ weights for the task and only adapt them during the finetuning process.

\subsection{SNIP Based Sparse Mask (SHiRA-SNIP)} 
SNIP \cite{lee2018snip} combines both weight and gradient based schemes and is computed as the magnitude of the product of the weight and its corresponding gradient. This formulation effectively captures the interplay between the weight magnitude, which reflects its overall contribution to the model's output, and its gradient information, which encodes the weight's task-specific relevance during finetuning. 

SNIP for a weight parameter is defined as:
\begin{equation}
      SNIP \triangleq |\langle\Theta_{i},\nabla_{\theta_i}\mathcal{L}\rangle|
\end{equation}
where $\langle. \rangle$ represents inner product, $\Theta_{i}$ is the weight parameter, $\nabla_{\theta_i}\mathcal{L}$ is the gradient of weight parameter with respect to the task loss $\mathcal{L}$ for the $i$th parameter in the network. 

\vspace{0.5in}
\section{Fuse and Scatter Op implementation}\label{sec::scatterop}
\begin{figure}[h]
    \centering
    \includegraphics[width=0.7\linewidth]{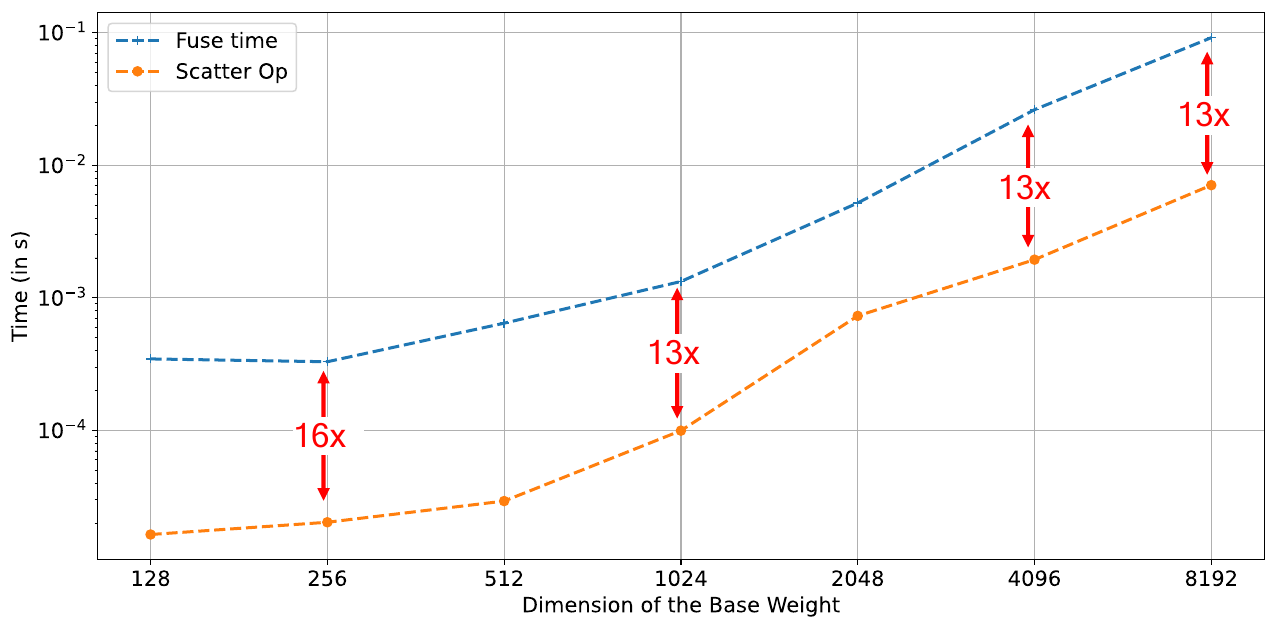}\vspace{-3mm}
    \caption{Comparison between average times for LoRA-fuse and SHiRA-\texttt{scatter\_op} implementation for 50 randomly initialized weights of various dimensions on a CPU (e.g., dimension = $4096$ means that the weight has shape $4096\times 4096$). For fusing, we compute time taken to merge LoRA adapters into the base weights (W + AB). Similarly, for the \texttt{scatter\_op}, we report time taken to overwrite base weights with SHiRA weights using the scatter op (\texttt{torch.Tensor.scatter\_}) based implementation in Pytorch.}
    \label{fig:speed}
\end{figure}

In this section, we compare fusing times of LoRA with our efficient \texttt{scatter\_op} (\texttt{torch.Tensor.scatter\_}) based implementation for SHiRA. For our experiments, we perform benchmarking on a Desktop-grade CPU and compute the average times for various tensor dimensions (e.g., tensor dimension = 4096 implies a weight of size $4096\times4096$, which is typical in modern LLMs). As shown in Fig.~\ref{fig:speed}, our \texttt{scatter\_op}-based SHiRA inference pipeline is up to $13\times$-$16\times$ faster than fusing LoRA weights, specially for larger dimensions.

Next, we present end-to-end switching times for prevalent LVMs and LLMs: SDXL and LLaMA2-7B. Notably, even for a smaller model like SDXL (2.6B params compared to 7B params in LLaMA2-7B), SHiRA achieves a 4.68x faster switching time (0.77s vs. 3.6s), while for LLaMA2-7B, with larger tensor dimensions, SHiRA attains a 5.71x speedup (4.93s vs. 28.15s) on a consumer grade CPU (see Table~\ref{tab:E2E-time}). Note that, fusing LoRA adapters for LLaMA2-7B on a CPU is 28.15s (nearly half a minute). Indeed, waiting half a minute for the adapter to switch/fuse is quite substantial and hampers user experience significantly. In contrast, SHiRA can get the adapter ready for inference within 4.93s, thus significantly improving the user experience. Note that, once the adapters are fused, inference time on the hardware is equal for both LoRA and SHiRA. Moreover, as discussed in~\cite{hflora}, for unfused LoRA case (which can enable rapid switching), the inference latency can be up to 30\% higher which is not the case with SHiRA. 

\begin{table}[]
    \centering
    \scalebox{0.95}{
    \begin{tabular} {  lccccc }
    \toprule
    \textbf{Model} & \textbf{LoRA} & \textbf{SHiRA} & \textbf{Speed-up} \\\hline
    \midrule
    \texttt{SDXL} &  $3.64 \pm 0.10$ & $\bm{0.77 \pm 0.09}$ & {\color{ForestGreen}$\bm{4.68\times}$}\\
    \texttt{LLaMA2-7B} & $28.15 \pm 1.62$ &  $\bm{4.93 \pm 0.23}$ & {\color{ForestGreen}$\bm{5.71\times}$} \\
    \bottomrule \\
\end{tabular}
}
    \caption{End-to-End switching time on CPU for SDXL and LLaMA2-7B: We achieve a very high ($4.7\times$-$5.7\times$) speed up in switching time compared to LoRA.}
    \label{tab:E2E-time}
\end{table}

\section{Turn any Trainer into SHiRA: Gradient Hook based Implementation}\label{Latencyeff}

In this section, we provide a method to convert any floating point training into SHiRA based finetuning. Specifically, SHiRA can be implemented directly using a functionality called 
\texttt{post\_accumulate\_gradient\_hooks} available in Pytorch 2.1.0. This \texttt{gradient\_hook} can be used to mask gradients after the gradient accumulation step is completed. Moreover, this enables us to apply SHiRA on any publicly available trainer (e.g., \texttt{Transformers.Trainer, SFT\_Trainer}, etc.). Therefore, implementing SHiRA on any task is trivial and can be done even without PEFT library, thus making SHiRA very easy to implement.

With this gradient hook based implementation, we were able to train all our adapters (including for models such as LLaMA-7B, LLaMA2-7B and SD-1.5) on a single NVIDIA A100 GPU at nearly the same speed as PEFT based LoRA implementation. SHiRA runs at 2.17 it/sec as compared to LoRA which is at 2.42 it/sec for LLaMA-7B finetuning.

\section{Latency- and Memory-Efficient PEFT based Implementation for SHiRA}\label{Memoryeff}
As discussed in Appendix~\ref{sec::scatterop},  \texttt{scatter\_op} can be utilized to manage sparse weight updates during inference. Given that SHiRA only finetunes a small subset of the pretrained model weights, we adopt a similar \texttt{scatter\_op}-based approach for training. This allows us to retain only the sparse training parameters in the optimizer, thereby significantly reducing the peak GPU memory utilization during training. As shown in Table~\ref{tab:memoryshira}, SHiRA not only trains at almost similar speed as LoRA, but also consumes $\sim16\%$ lower  peak GPU memory. Compared to other variants like DoRA, SHiRA training consumes significantly ($\sim40\%$) lower peak GPU memory and also trains much faster (SHiRA is about 36\% faster than DoRA). All memory requirement data was collected using \texttt{psutil} utility used within the \texttt{Transformers.Trainer} training loop for LLaMA2-7B.

Finally, note that, partial finetuning techniques proposed in the pre-LoRA era~\cite{zhao2020masking, sung2021training, ansell2021composable, xu2021raise, guo2020parameter} do not have such memory-efficient implementations, which makes them impractical for large generative models. Therefore, SHiRA significantly outperforms prior partial finetuning techniques in training memory costs and is highly practical for modern LVM and LLM adaptations tasks.
\begin{table}[h]
    \addtolength{\tabcolsep}{-2pt}
    \centering
    %\fontsize{6.75pt}{6.75pt}\selectfont
    
    \begin{tabular} {  lccc }
    \toprule
    \centering \textbf{Adapter} & \textbf{Peak GPU memory (GB)} & \textbf{\textit{\#}Training steps/s} \\\hline
    \midrule
    \texttt{LoRA-PEFT} &  $35.10$ & $0.69$\\
    \texttt{DoRA-PEFT} & $49.49$ (\color{red} +40.99$\%$) & $0.49$ (\color{red} -28.98\%)\\
    \texttt{SHiRA-PEFT} & $29.26$ (\color{ForestGreen} -16.63\%) &  $0.67$ (\color{red} -2.89\%)\\
    \bottomrule
\end{tabular}\vspace{1mm}
    \caption{Peak GPU memory consumption (in GBs) and \textit{\#}Training steps per second during training for PEFT-based implementation of various adapters for LLaMA2-7B. Relative changes compared to LoRA are highlighted: {\color{ForestGreen} Green} indicates improved performance (lower memory consumption, faster training speed), while {\color{red} Red} indicates degraded performance (higher memory consumption, slower training speed). SHiRA trains at nearly the same speed as LoRA but consumes up to 16\% lower peak GPU memory.} 
    \label{tab:memoryshira}
\end{table}

\section{Proofs of Lemma}\label{lemma-proof}

\subsection{Lemma~\ref{lemmaComp}}\label{prooflemmaComp}
\textbf{Lemma 4.1.} \textit{The parameter complexity and learning complexity of SHiRA is equal to the number of non-zero elements in the adapter.}
\begin{proof}
The parameter complexity and learning complexity depends on the parameters to be learned. The number parameters of the adapter is equal to $\mid \mid \Delta \mathbf{W} \mid \mid_{0}$.
\end{proof}

\subsection{Lemma~\ref{lemmaSing}}\label{prooflemmaSing}
\textbf{Lemma 4.2.} %For any given SHiRA adapter of size $(m,n)$ with sparsity ratio $\rho$ we can construct an $r$ rank LoRA with approximation error bounded by $\sigma_{r+1}^2$.
\textit{If we specify a sparsity factor, the LoRA is $r$ rank approximation of SHiRA with approximation error bounded by $\sigma_{r+1}^2$, the $(r+1)^{th}$ singular value of the SHiRA adapter.}
\begin{proof}
Let $\Delta\mathbf{W}$ be the given SHiRA adapter of size $(m, n)$ and  sparsity factor $\rho$. Consider the SVD decomposition of $\Delta \mathbf{W}$. Next, we construct an $r$ rank matrix
approximation using the $r$ largest singular values of the adapter. This reconstructed $r$ rank matrix can be seen as a LoRA adapter. Based on Eckart-Young theorem~(\cite{eckart1936approximation}) and theorem 4.95 in~\cite{deisenroth2020mathematics}, the approximation error is equal to $(r+1)$-th singular value of the SHiRA adapter ($\sigma_{r+1}^{2}$). If the  $\Delta \mathbf{W}$ is an $r$ rank matrix then the approximation error is zero.
\end{proof}

\subsection{Lemma~\ref{lemmaScale}}\label{prooflemmaScale}
\textbf{Lemma 4.3.} \textit{Scaling factor for SHiRA is independent of the rank of the adapter and can be set to 1.}
\begin{proof}
The LoRA update equation for any given adapter is as follows:
\begin{equation}
    \mathbf{Y_{out}} = (\mathbf{W} + \alpha_r \mathbf{BA})\mathbf{X_{in}} + \mathbf{b}. 
\end{equation}
Note $\alpha_r = \frac{\alpha}{r}$ is the scaling factor, where $\alpha$ is a hyperparameter and $r$ is the rank. Three possible initialization for $\mathbf{A}$ and $\mathbf{B}$ are as follows:
\begin{itemize}
\item if $A$ and $B$ are initialized to zero, no learning occurs since this corresponds to saddle point~\cite{hayou2024lora+}. 
\item $A$ and $B$ are  initialized to $\mathcal{N}(0,\sigma_{a}^2)$ and $0$ respectively. Here, $\sigma_{a}^2 = \Theta(n^{-1})$, to ensure that $A^Tx_{i}$ remains bounded with width $n$  of the adapter.
\item $A$ and $B$ are  initialized to $0$ and $\mathcal{N}(0,1)$ respectively. Here, it is important to note that the variance of $B$ does not depend of the width of the adapter.
\end{itemize}
However, to avoid gradient collapse for higher ranks,~\cite{rslora} recommends to set $\alpha_r$ as $\frac{\alpha}{\sqrt{r}}$. Further, optimal convergence the update of $A$ and $B$ matrix updates have 
different learning rates \cite{hayou2024lora+}.
For the SHiRA adapter, the update equation is given below:
\begin{equation}
    \mathbf{Y_{out}} = \mathbf{(W + S)X_{in} + b}. 
\end{equation}
where, $S$ is the sparse matrix with a designed sparsity ratio. All non-zero locations in $S$ are \textit{implicitly} initialized to the base matrix weights. This initialization ensures that the updates remain
bounded during the finetuning  stage using stochastic gradient descent. It is also important to note that the scaling is independent of the rank for SHiRA.
\end{proof}

\subsection{Lemma~\ref{lemma-null}}\label{proof-lemma-null}
\textbf{Lemma 4.4.} \textit{Consider two adapters, $\Delta W_{1}$ and $\Delta W_{2}$. If one of the adapters, $\Delta {W_{1}}$ or $\Delta {W_{2}}$ lies in the null space of the other, then the  adapters will not interfere multiplicatively.}
\begin{proof}
The proof leverages two facts: (\textit{i})~$\Delta {W_{1}}^{T}\Delta {W_{2}} = \mathbb{O}$ given that one adapter lies in the null space of other. Here, $\mathbb{O}$ is a zero matrix ($\mathbb{O}_{i,j}=\{0\} \forall i, j$). (\textit{ii})~Power series expansion of the non-linear activation function: The power series expansion has terms involving the matrix product of adapters. Since each adapter is in the null space of the other, all terms involving product of adapters are equal to zero. Therefore the adapters do not interfere multiplicatively.
\end{proof}

This lemma can be  extended to a scenario with more than two parallel additive  adapters. If all possible pairs of adapters lie in the null space of each others all cross-terms between adapters are zero.

\subsection{Lemma \ref{lemma-struct}} \label{proof-lemma-struct}
\textbf{Lemma 4.5.} \textit{Non-overlapping SHiRA-Struct adapters are nearly orthogonal. That is, AWOR for non-overlapping SHiRA-Struct adapters is at most the sum of sparsity of individual adapters. Since all SHiRA masks are highly sparse, this means that the product $\mathcal{A}_1^T\mathcal{A}_2$ has a lot of zeros, thus making the adapters nearly orthogonal.}

\begin{proof}
Continuing from the adapter definitions used in the main text for this lemma, let us compute $\mathcal{A}_1^T\mathcal{A}_2$ and then analyze its AWOR:
\begin{equation}
\begin{aligned}
    \mathcal{A}_1^T\mathcal{A}_2& = (\mathbb{I} + \mathcal{S}_1)^T(\mathbb{I} + \mathcal{S}_2) = \mathbb{I} + \mathbb{I}\mathcal{S}_2 + \mathcal{S}_1^T\mathbb{I} + \mathcal{S}_1^T\mathcal{S}_2 = \mathbb{I} + \mathcal{S}_2 + \mathcal{S}_1^T
    % \mathcal{A}_1^T\mathcal{A}_2& = (\mathbb{I} + \mathcal{S}_1)^T(\mathbb{I} + \mathcal{S}_2)\\
    % &= \mathbb{I} + \mathbb{I}\mathcal{S}_2 + \mathcal{S}_1^T\mathbb{I} + \mathcal{S}_1^T\mathcal{S}_2\\
    % &= \mathbb{I} + \mathcal{S}_2 + \mathcal{S}_1^T
\end{aligned}
\label{EqAworStruct}
\end{equation}

Here, $\mathcal{S}_1^T\mathcal{S}_2$ is zero by design because $\mathcal{S}_1$ and $\mathcal{S}_2$ do not have common non-zero rows. Moreover, since both $\mathcal{S}_1$ and $\mathcal{S}_2$ are highly sparse, $\mathcal{A}_1^T\mathcal{A}_2$ has a sparsity equal to the sum of sparsity of $\mathbb{I}$, $\mathcal{S}_1$ and $\mathcal{S}_2$. Note that, $\mathbb{I}+\mathcal{S}_2=\mathcal{A}_2$. Thus, AWOR for non-overlapping SHiRA-Struct adapters is at most the sum of sparsity of individual adapters.
\end{proof}

\section{Dataset and Evaluation Metric Descriptions}\label{dataset-desp}
\subsection{Datasets}
\subsubsection{Language Datasets}
\begin{wraptable}{r}{5.0cm}
    \vspace{-0.7cm}
% \begin{table*}[h]
    \addtolength{\tabcolsep}{-2pt}
    \centering
    \fontsize{7.0pt}{5.75pt}\selectfont
    % \fontsize{8.00pt}{8.25pt}\selectfont
    \begin{tabular} {  lccc }
    \toprule
    \textbf{Dataset} & \textbf{\#Train} & \textbf{\#Val} & \textbf{Test} \\\hline
    \midrule
    PiQA & 16K & 2K & 3K \\
    \midrule
    BoolQ & 9.4K & 2.4K &  2.4K \\
    \midrule
    SIQA & 33.4K & 1.9K &  1.9K \\
    \midrule
    OBQA & 4.9K & 0.5K &  0.5K\\
    \midrule
    Winogrande & 9.2K & 1.3K &  1.8K\\
    \midrule
    HellaSwag & 39.9K & 10K & 10K \\
    \midrule
    Arc\_easy & 2.25K & 570 & 2.36K \\
    \midrule
    Arc\_challenge & 1.12K & 299 & 1.12K \\
    \bottomrule
\end{tabular}
    \vspace{-1mm}
    \caption{Commonsense Benchmarks} 
    \label{tab: data_commonsense}
% \end{table*}
\vspace{-5mm}
\end{wraptable}
For language finetuning tasks, we use the commonsense reasoning datasets, which comprise 8 sub-tasks, each with a predefined training and testing set as shown in Table \ref{tab: data_commonsense}. We follow the setting of \cite{llmadapters} for SHiRA Single Adapter training. The common sense reasoning training dataset is a combination of the training datasets provided by \cite{hudson2019gqa}, while we evaluate each evaluation dataset separately as in Table \ref{tab: CommonSense-Single-Adapter}. For multi-adapter LLM experiments, we train each adapter from one particular task, and then perform multi-adapter evaluation on all the tasks.

\subsubsection{Vision Datasets}\label{visiondataset}
For style transfer adaptation tasks as described in sections~\ref{single-shira-lvm} and \ref{lvm-2}, we use two datasets, Bluefire and Paintings. Images present in both of these datasets are collected from public-domain (CC-0 license). 

The Bluefire dataset consists of a total of 54 images consisting of 6 different concepts - Cars, Dragons, Birds, Foxes, Men and Castles. For all these concepts, images with "blue-fire" effect are collected and used for style transfer finetuning. The validation of the Bluefire dataset consists of 30 images. 9 of the 30 images contain one of the 6 concepts in the training set, and the rest 21 are new. A few examples of unseen concepts in the validation set: \textit{football, monster, sword, chess rook, lion, koala etc}.

Similarly, the painting datasets contain a total of 90 images of "painting" style images of 9 different concepts - fire, birds, elephants, ships, horses, flowers, women, men and tigers. The validation set of the Paintings dataset consists of 21 images, out of which 9 contain concepts from the training set. The remaining 12 are new concepts not included in the training set. A few examples of unseen concepts in the validation set: \textit{lion, tiger, dog, cat, koala, panda, and other landscapes}.

\subsection{Evaluation Metrics} 
\paragraph{HPSv2 metric evaluation} 
For all style transfer finetuning experiments with Bluefire and Paintings dataset, we report HPS metric to quantify the quality of the generated images. For Bluefire validation, 30 images per validation prompt are generated for different seeds, hence generating 900 images for HPS analysis. We follow a similar paradigm for Paintings and generate 630 images with 21 prompts.

\section{Training Details}\label{training-details}
In this section, we list hyperparameters used for our experiments for Language and Vision finetuning tasks in Table \ref{tab: train_details}.

\begin{table*}[h]
    \centering
    \fontsize{7.0pt}{5.75pt}\selectfont
    \begin{tabular} {  lcccccccc }
    \toprule
    
    \multirow{1}{*}{\textbf{Method}} & \multirow{1}{*}{\textbf{Adapter Target Modules}} & \multirow{1}{*}{\textbf{Optimizer}} &  \multirow{1}{*}{\textbf{LR}} & \multirow{1}{*}{\textbf{LR-Scheduler}} & \multirow{1}{*}{Rank}\\
    \hline
    \midrule

    LoRA LVM & \multirow{5}{*}{q-proj,k-proj,v-proj,up-proj,down-proj} & \multirow{5}{*}{AdamW} & $1e-4$ & Cosine & 64\\
    SHiRA LVM &  &  & $1e-4$ & Cosine & NA\\
    LoRA LLM &  &  & $2e-4$ & Linear & 32\\
    DoRA LLM &  &  & $2e-4$ & Linear & 32 \\
    SHiRA LLM & &  & $5e-4$ & Linear & NA \\
    % Multi-LoRA LLM &  &  & $4e-4$ & Linear & 64\\
    % Mutli-SHiRA LLM & &  & $4e-4$ & Linear & NA \\
    
    \midrule
    \end{tabular}\vspace{-1mm}
    \caption{Training hyperparameters used for finetuning experiments.} 
\label{tab: train_details}
\end{table*}
\label{tab: detailsllm-multi}

 All finetuning and evaluation experiments for language and vision tasks are done using a single NVIDIA A100 GPU.

\section{Effect of Scaling Factor $\alpha$ during Inference}\label{alpha-effect}
\begin{figure}[t]
  \centering
   \includegraphics[width=1.0\linewidth]{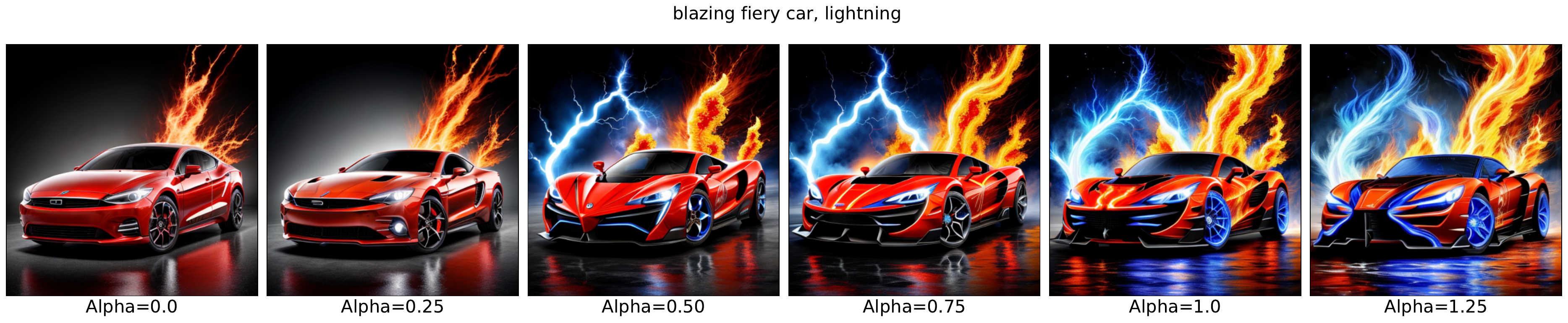}
   \caption{Effect of $\alpha$ scaling on image quality. $\alpha=0.0$ is the base model output without any adapter effects. We can see that as the $\alpha$ increases, the SHiRA adapter effect increases similar to how it works for LoRA inference. }
   \label{fig:alphaFigure}
\end{figure}

As described in section~\ref{shira-approach}, in order to adapt the pretrained model to a new task, we only finetune very few weight parameters relevant to the task. For our adapter, we can easily extract out these modified weights as $S = W_{new} - W$, where $W_{new}$ is the weight obtained after SHiRA training, and $W$ is the prertained weight. Since only $1$-$2\%$ weights change during SHiRA training, $S$ is highly sparse and thus constitutes our sparse adapter. Hence, the new finetuned weights of the base model can be viewed as $W_{new} = W + S$.%, where $W_{new}$ is the weight after finetuning, $W$ is the pretrained weights and $S$ is our sparse adapter. 

Similar to LoRA, the strength of SHiRA adapter at inference time can be modified using a scaling factor $\alpha$. For any defined $\alpha$ scaling, the new weights of the model can be expressed as $W_{new} = W + \alpha S$. Fig.~\ref{fig:alphaFigure} shows the effect of varying $\alpha$ on the output image. As evident, choosing an  $\alpha < 1$ reduces the "blue fire" in the generated image and whereas $\alpha > 1$ amplifies the style transfer effect. For $\alpha=0.0$, the adapter is disabled and the model's output is the same as that for the base model.

\section{More Detailed Comparison among Various Masks}\label{tab11}
We provide HPSv2 scores for all SHiRA masking schemes in Table~\ref{tab: t2i_table}.
\begin{table*}[h]
    %\addtolength{\tabcolsep}{-2pt}
    \centering
    \fontsize{7.0pt}{5.75pt}\selectfont
    % \fontsize{8.00pt}{8.25pt}\selectfont
    \begin{tabular} {  lccccccccc }
    \toprule
    
    \multirow{1}{*}{\textbf{Adapter Style}} & \multirow{1}{*}{\textbf{Adapter Method}} & \textbf{\%Params} &  \multicolumn{3}{c}{\textbf{HPSv2 score($\uparrow$)}} \\

    & & &  $\alpha=1$ & $\alpha=0.75$ & $\alpha=0.5$ \\
    
    \hline
    \midrule
    
    %\toprule
    \multirow{5}{*}{\vspace{-0.15in}Paintings} & LoRA & 3.84 & $24.7 \pm1.8$ & $28.4 \pm 1.4$ & $31.3 \pm 1.5$ \\
     & SHiRA-Struct &  \textbf{1.99}& $\bm{31.2 \pm 1.7}$ & $\bm{32.1 \pm 1.8}$ & $\bm{33.0 \pm 1.8}$ \\
    & SHiRA-Rand & 2.05& $30.7 \pm 1.9$ & $31.7 \pm 1.8$ & $32.7 \pm 1.9$ \\
    & SHiRA-WM & 2.05 & $29.7 \pm 1.9$ & $30.6 \pm 1.7$ & $32.1 \pm 1.8$ \\
    & SHiRA-Grad & 2.05 & $30.3 \pm 1.8$ & $31.3 \pm 1.7$ & $32.3 \pm 1.8$ \\
    & SHiRA-SNIP & 2.05 & $29.8 \pm 1.8$ & $30.8 \pm 1.8$ & $31.6 \pm 1.8$ \\
    \midrule
    \multirow{5}{*}{\vspace{-0.15in}Bluefire} &  LoRA & 3.84 & $32.6 \pm 1.9$ & $34.1 \pm 1.5$ & $33.6 \pm 1.6$ \\
    & SHiRA-Struct & \textbf{1.99} & $\bm{34.2 \pm 1.6}$ & $\bm{34.7 \pm 1.5}$ & $\bm{34.1 \pm 1.5}$  \\
    & SHiRA-Rand & 2.05 & $33.4 \pm 1.9$ & $34.1 \pm 1.5$ & $33.7 \pm 1.7$ \\
    & SHiRA-WM & 2.05 & $31.9 \pm 2.1$ & $33.3 \pm 1.6$ & $33.1 \pm 1.7$ \\
    & SHiRA-Grad & 2.05 & $\bm{34.2 \pm 1.5}$ & $34.4 \pm 1.5$ & $33.7 \pm 1.7$ \\
    & SHiRA-SNIP & 2.05 & $33.7 \pm 1.7$ & $34.3 \pm 1.4$ & $33.7 \pm 1.6$ \\
    
    \bottomrule
\end{tabular}
    \caption{Comparison between LoRA and various SHiRA schemes with respect to HPSv2 metric. For vision problems, SHiRA-Struct outperforms all other methods.} 
    \label{tab: t2i_table}
\end{table*}

\section{More Results}\label{AppMoreRes}

\subsection{Additional Sample Images for Vision Style Transfer Applications}\label{AppMoreSamplesLVM}
\begin{figure}[h!]
  \centering
   \includegraphics[width=1.0\linewidth]{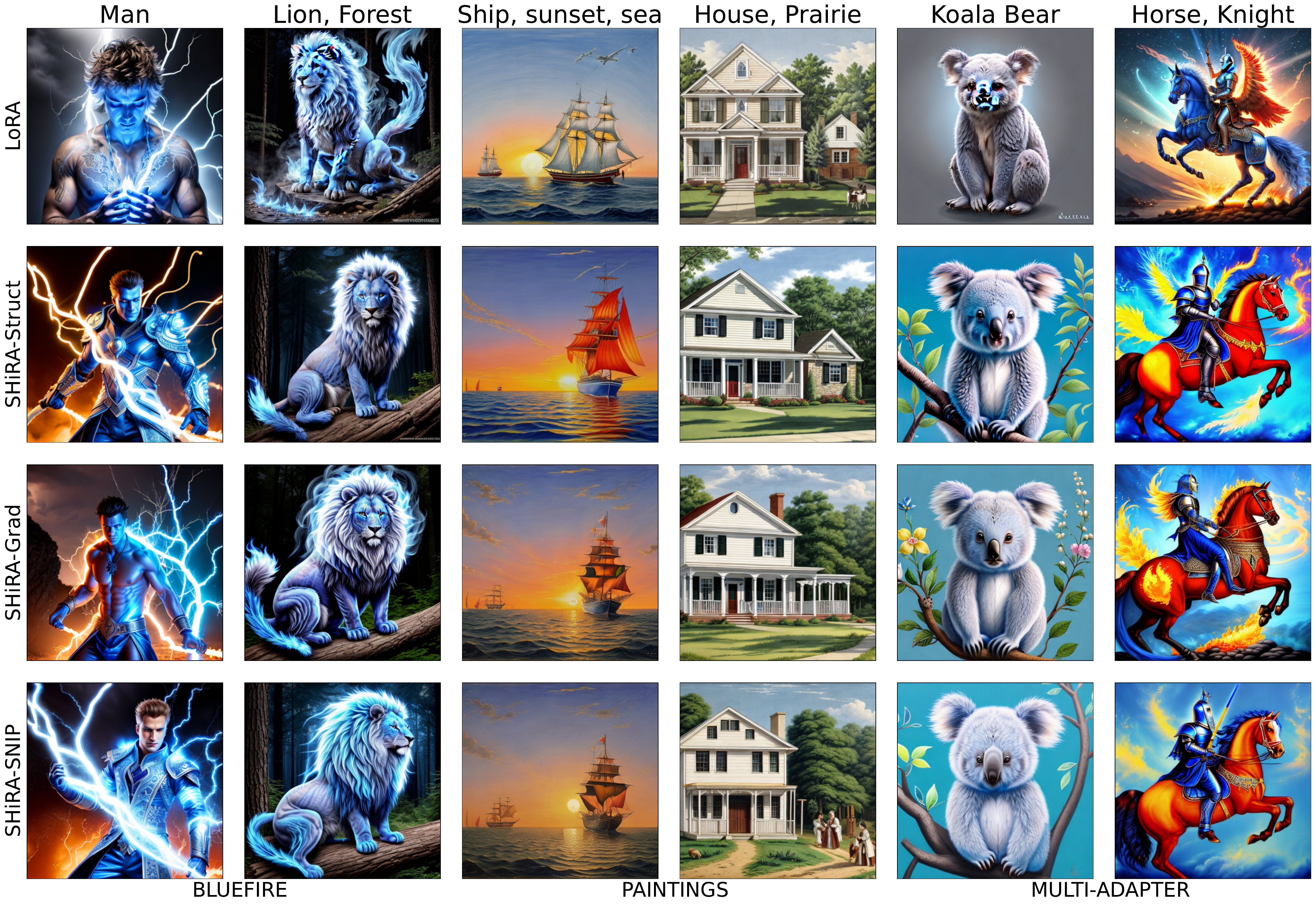}
   \caption{More image samples for single and multi-adapter fusion. We observe that LoRA exhibits artifacts for koala and concept loss for knight in Multi-Adapter fusion while SHiRA produces significantly better images.}
   \label{fig:appendixfig}
\end{figure}

\begin{figure}[h!]
  \centering
   \includegraphics[width=1.0\linewidth]{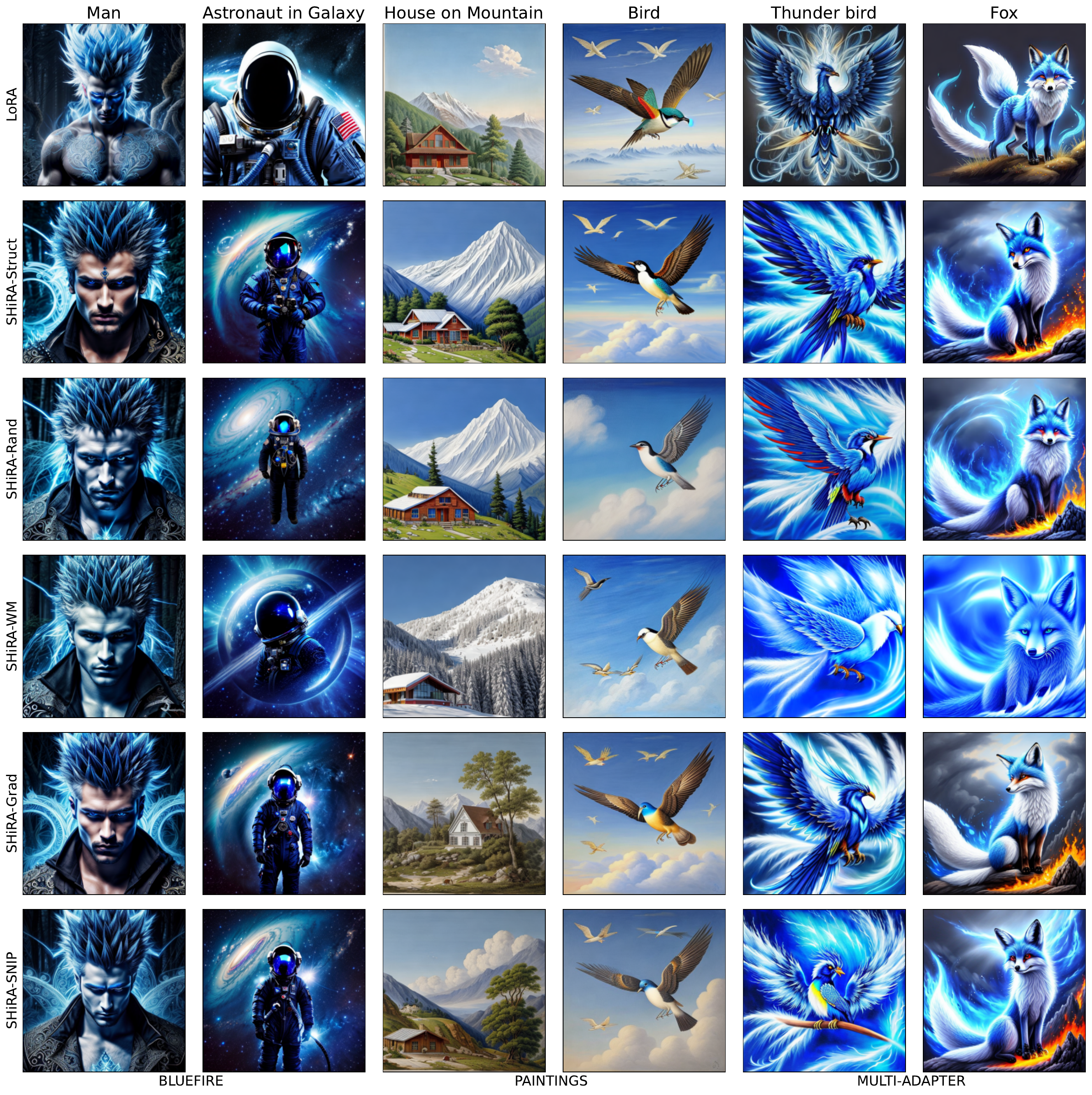}
   \caption{More image samples for single and multi-adapter fusion. We observe that LoRA images exhibit concept loss for bird in Multi-Adapter fusion.}
   \label{fig:shiraFigure2}
\end{figure}

\begin{figure}[h!]
  \centering
   \includegraphics[width=1.0\linewidth]{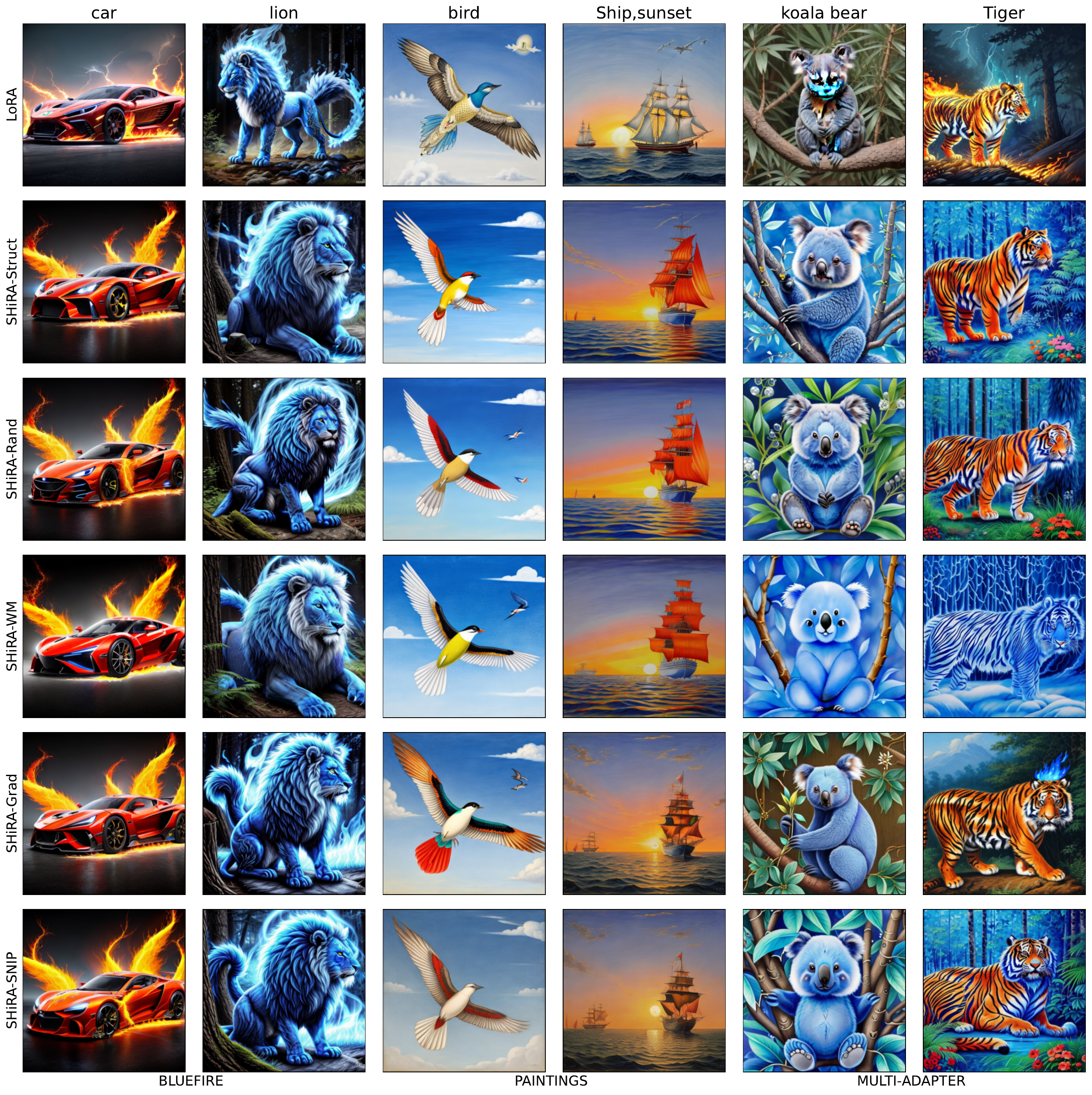}
   \caption{More image samples for single and multi-adapter fusion. Koala is not included in the training set of either of the Bluefire and Paintings Adapter styles. We observe that for this class, LoRA has significant artifacts whereas SHiRA produces exceptional images.}
   \label{fig:shiraFigure3}
\end{figure}

\begin{figure}[h!]
  \centering
   \includegraphics[width=1.0\linewidth]{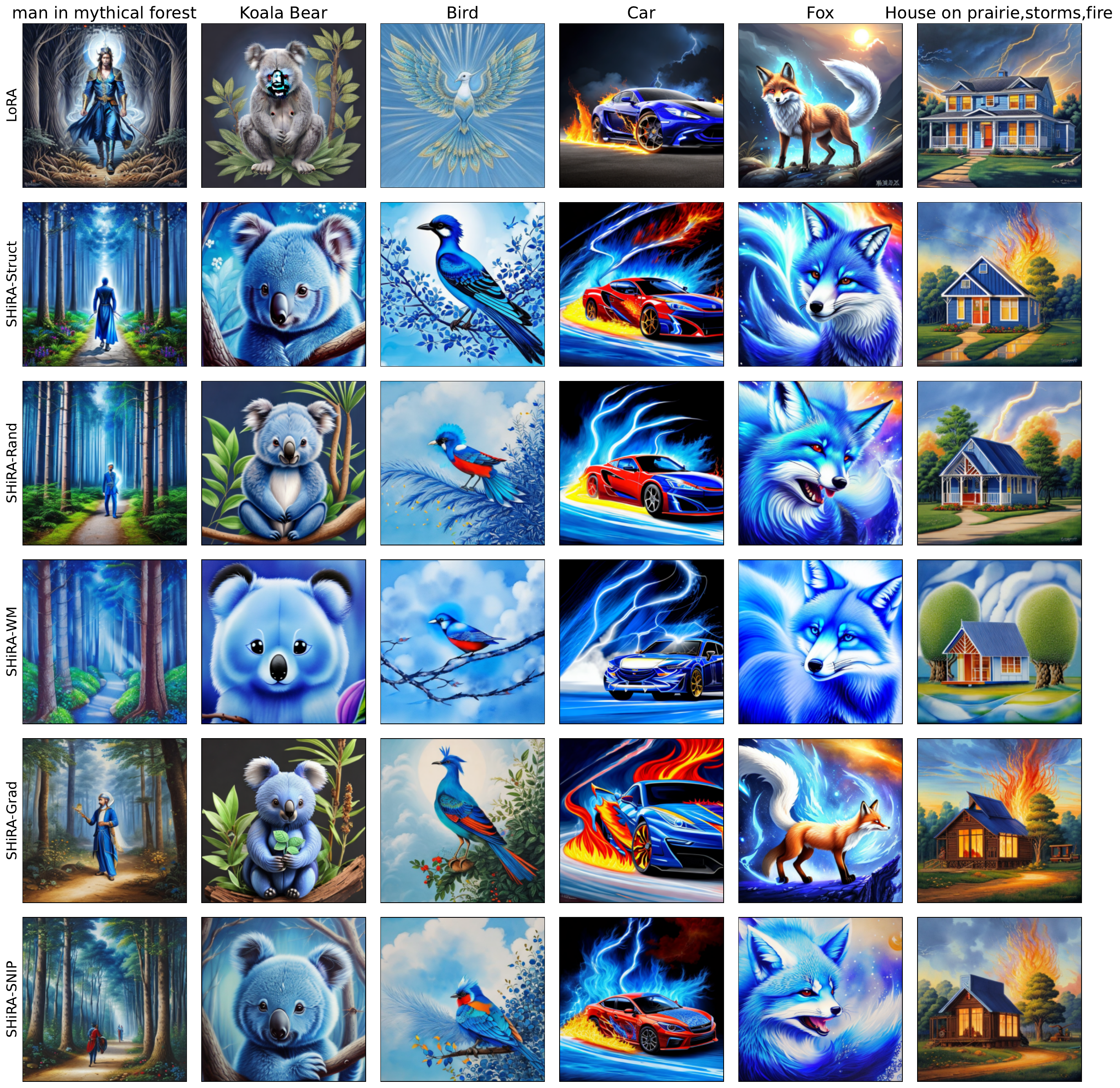}
   \caption{More results for multi-adapter fusion. Koala is not included in the training set of either of the Bluefire and Paintings Adapter styles. We observe that for this class, LoRA has significant artifacts whereas SHiRA produces exceptional images.}
   \label{fig:shiraFigure4}
\end{figure}

We show many more sample images for various adaptation usecases in Fig. ~\ref{fig:appendixfig}, ~\ref{fig:shiraFigure2}, ~\ref{fig:shiraFigure3}, and~\ref{fig:shiraFigure4}.

\subsection{Image Classification and GLUE}
\begin{table}[t]
    \centering
    %\fontsize{6.75pt}{6.75pt}\selectfont
    
    \begin{tabular} {  lccccc }
    \toprule
    \centering \textbf{Adapter} & \textbf{cifar10} & \textbf{cifar100}  & \textbf{food101}  & \textbf{dtd} \\\hline
    \midrule
    \texttt{LoRA} &  $97.94$ & $87.97$ & $84.27$ & $69.41$\\
    \texttt{SHiRA} & \textbf{98.05} &  \textbf{88.15} &  \textbf{84.43} &  \textbf{69.73} \\
    \bottomrule \\
\end{tabular}
    %\captionsetup{font=footnotesize}
    \caption{LoRA vs SHiRA for Image Classification using ViT-Base model. SHiRA consistently outperforms LoRA on these transfer learning tasks.}
    \label{tab:ViT}
\end{table}

\begin{table}[t]
    \centering
    \scalebox{0.75}{
    \begin{tabular} { lcccccc}
    \toprule
    \textbf{Adapter} & \textit{\#}\textbf{Params} & \textbf{COLA} & \textbf{QNLI} & \textbf{MPRC} & \textbf{SST2} & \textbf{Average}\\\hline
    \midrule
    \texttt{LoRA} & 1.33M & $69.73$ & $93.76$ & $89.71$ & $95.57$ & $87.19$ ({\color{red} +0\%})\\
    \texttt{SoRA} &  910K & \textbf{71.48} & \textbf{94.28} & $91.98$ & $95.64$ & \textbf{88.34} ({\color{ForestGreen} +1.15\%})\\
    \texttt{SHiRA} & \textbf{636K} & $70.62$ &  $93.90$ &  \textbf{92.15} &  \textbf{96.50} & \textbf{88.29} ({\color{ForestGreen} +1.10\%})\\
    \bottomrule \\
\end{tabular}
}
    \caption{GLUE benchmarking for the DeBERTa-V3-base.
As evident, with nearly 2x smaller adapter, SHiRA outperforms LoRA by 1.1\% accuracy on average. Further, SHiRA achieves a similar accuracy as SoRA while being 30\% smaller in adapter size. Hence, SHiRA generalizes to other language tasks as well.}
    \label{tab:GLUE}
\end{table}

We further conduct more experiments on image classification and GLUE tasks using SHiRA-WM. For image classification, we finetune Vision Transformer (ViT) using LoRA and SHiRA for four common transfer learning datasets, namely, CIFAR-10, CIFAR-100, Food101, and Describable Textures Dataset (DTD) (see Table~\ref{tab:ViT}). Both methods have comparable parameters around 300K. As shown in Table~\ref{tab:ViT}, we outperform LoRA on all image classification tasks.

For GLUE, we use the code released by SoRA~\cite{ding2023sparse} which relies on dynamically adjusting the ranks of the adapters. In Table~\ref{tab:GLUE}, we report accuracy on four common GLUE tasks: QNLI, COLA, SST2, and MRPC. Accuracy numbers for LoRA and SoRA are directly taken from the SoRA paper since we are using the official code to run SHiRA experiments. As evident, with nearly 2x smaller adapter, SHiRA outperforms LoRA by 1.1\% accuracy on average. Further, SHiRA achieves a similar accuracy as SoRA while being 30\% smaller in adapter size. Indeed, SoRA cannot enable rapid switching like SHiRA. Therefore, we again demonstrate that a simple approach like SHiRA-WM outperforms LoRA and its advanced variants with a similar or significantly better accuracy while providing additional deployment benefits.

\subsection{Analysis of Trained Adapters}\label{sec:analysis-adapters}
\begin{table}[t]
    \centering
    \scalebox{0.95}{
    %\fontsize{6.75pt}{6.75pt}\selectfont
    \begin{tabular} {lcccc}
    \toprule
    & \textbf{Base} & \textbf{Arc\_e} & \textbf{BoolQ} & \textbf{PIQA} \\\hline
    \midrule
    \textbf{Base} &  $0$ & $\bm{37.0}$ & $\bm{67.0}$ & $\bm{75.0}$ \\
    \textbf{Arc\_e} && $0$ & $75.0$ & $81.5$\\
    \textbf{BoolQ} &   &  & $0$ & $98.5$\\
    \textbf{PIQA} & &  &  & $0$\\
    \bottomrule \\
\end{tabular}
}
    %\captionsetup{font=footnotesize}
    \caption{L2 distances between pretrained base weights and SHiRA adapters vs. distances between adapters: Adapters are closer to the base model weights than to each other.} 
    \label{tab:L2}
\end{table}

\begin{figure}[h!]
\centering
\includegraphics[width=0.65\textwidth]{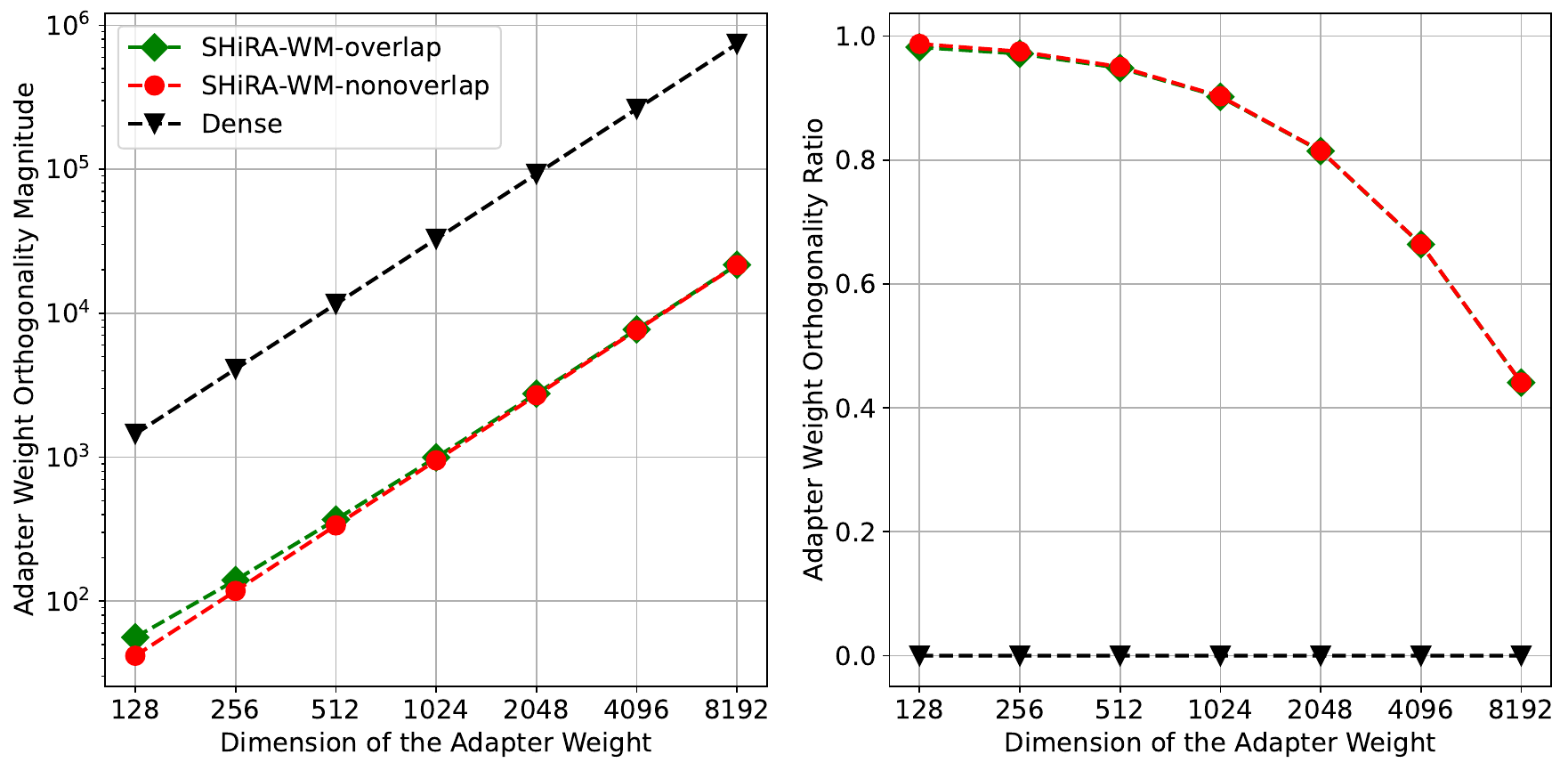}
    %\captionsetup{font=footnotesize}
    \caption{\fontsize{10pt}{12pt}\selectfont Adapter Weight Orthogonality Magnitude (AWOM: L2 magnitude) and Adapter Weight Orthogonality Ratio (AWOR: Sparsity Ratio) of the product $\mathcal{A}_1^T\mathcal{A}_2$ between two adapters for unstructured SHiRA-WM overlap and non-overlapping cases ($99\%$ sparse). We vary the adapter dimensions (e.g., $4096$ refers to a pretrained weight of dimensions $4096\times 4096$) and measure AWOM and AWOR for each weight size (averaged over 50 seeds). For unstructured SHiRA masks, overlapping and non-overlapping adapters achieve \textit{coinciding} AWOR and AWOM, thus suggesting that their orthogonality properties are very similar due to high sparsity. This explains our multi-adapter LLM results in Table~\ref{tab: multishira}.} 
    \label{orthogonality-figure}
\end{figure}

\paragraph{Are adapter tasks sufficiently different?} Table~\ref{tab:L2} shows the L2 analysis for the adapters trained in Table~\ref{tab: multishira}. We compute the L2 distance between each adapter and the original pretrained weights (all adapters train top 1\% weights in the overlap setting) as well as the L2 distance between each adapter. Clearly, each adapter is closer to the pretrained weights compared to the other adapters. This demonstrates that the tasks are sufficiently different.

\paragraph{Why does SHiRA-WM-Overlap perform well?}
Next, as shown in Fig. \ref{orthogonality-figure}, for unstructured SHiRA masks, both overlapping and non-overlapping adapters have identical AWOR and AWOM values. This suggests that their orthogonality characteristics are quite similar due to the high sparsity. We hypothesize that this is the main reason for the good performance of SHiRA-WM-overlap and explains the results in Table~\ref{tab: multishira}.

\section{Societal Impact}\label{societal-impact}
Our work enables on-device deployment of adapters which can have a clear positive impact on society as it allows for privacy-preserving generative AI use. With our work, users would be able to rapidly generate images in specific styles directly on-device. On the other hand, while efficient finetuning techniques have many advantages, they bring the potential risk of digital forgery. This is mainly due to finetuning the generative models on a much smaller subset of data, leading to potential overfitting. As our proposed method is also a parameter-efficient finetuning technique, it suffers from similar potential risk as the other PEFT algorithms. 

\section{Limitations and Future Work}
In this work, we show that our proposed sparse high rank adapter, SHiRA, with merely finetuning 1-2\% parameters of the pretrained generative models is sufficient to achieve high performance on many adapter tasks. However, in order to adopt our method for mobile deployment, hardware-software co-design techniques, such as lookup-table (LUT) based approaches, may be necessary to optimize the implementation for edge devices. 

Moreover, as discussed in the main text, building optimal sparse masks (i.e., which parameters to train for a given task) warrants further investigation. 

\clearpage

%% file: sections/checklist.tex
\section*{NeurIPS Paper Checklist}

%%% END INSTRUCTIONS %%%

\begin{enumerate}

\item {\bf Claims}
    \item[] Question: Do the main claims made in the abstract and introduction accurately reflect the paper's contributions and scope?
    \item[] Answer: \answerYes{} % Replace by \answerYes{}, \answerNo{}, or \answerNA{}.
    \item[] Justification: The manuscript discusses and reports detailed results accurately reflecting the claims and the scope of the work.
    \item[] Guidelines:

    \begin{itemize}
        \item The answer NA means that the abstract and introduction do not include the claims made in the paper.
        \item The abstract and/or introduction should clearly state the claims made, including the contributions made in the paper and important assumptions and limitations. A No or NA answer to this question will not be perceived well by the reviewers. 
        \item The claims made should match theoretical and experimental results, and reflect how much the results can be expected to generalize to other settings. 
        \item It is fine to include aspirational goals as motivation as long as it is clear that these goals are not attained by the paper. 
    \end{itemize}

\item {\bf Limitations}
    \item[] Question: Does the paper discuss the limitations of the work performed by the authors?
    \item[] Answer: \answerYes{} % Replace by \answerYes{}, \answerNo{}, or \answerNA{}.
    \item[] Justification: Yes, discussed in section~\ref{sec:appendix}.
    \item[] Guidelines:

    \begin{itemize}
        \item The answer NA means that the paper has no limitation while the answer No means that the paper has limitations, but those are not discussed in the paper. 
        \item The authors are encouraged to create a separate "Limitations" section in their paper.
        \item The paper should point out any strong assumptions and how robust the results are to violations of these assumptions (e.g., independence assumptions, noiseless settings, model well-specification, asymptotic approximations only holding locally). The authors should reflect on how these assumptions might be violated in practice and what the implications would be.
        \item The authors should reflect on the scope of the claims made, e.g., if the approach was only tested on a few datasets or with a few runs. In general, empirical results often depend on implicit assumptions, which should be articulated.
        \item The authors should reflect on the factors that influence the performance of the approach. For example, a facial recognition algorithm may perform poorly when image resolution is low or images are taken in low lighting. Or a speech-to-text system might not be used reliably to provide closed captions for online lectures because it fails to handle technical jargon.
        \item The authors should discuss the computational efficiency of the proposed algorithms and how they scale with dataset size.
        \item If applicable, the authors should discuss possible limitations of their approach to address problems of privacy and fairness.
        \item While the authors might fear that complete honesty about limitations might be used by reviewers as grounds for rejection, a worse outcome might be that reviewers discover limitations that aren't acknowledged in the paper. The authors should use their best judgment and recognize that individual actions in favor of transparency play an important role in developing norms that preserve the integrity of the community. Reviewers will be specifically instructed to not penalize honesty concerning limitations.
    \end{itemize}

\item {\bf Theory Assumptions and Proofs}
    \item[] Question: For each theoretical result, does the paper provide the full set of assumptions and a complete (and correct) proof?
    \item[] Answer: \answerYes{} % Replace by \answerYes{}, \answerNo{}, or \answerNA{}.
    \item[] Justification: Discussed in section Appendix~\ref{lemma-proof}.
    \item[] Guidelines:

    \begin{itemize}
        \item The answer NA means that the paper does not include theoretical results. 
        \item All the theorems, formulas, and proofs in the paper should be numbered and cross-referenced.
        \item All assumptions should be clearly stated or referenced in the statement of any theorems.
        \item The proofs can either appear in the main paper or the supplemental material, but if they appear in the supplemental material, the authors are encouraged to provide a short proof sketch to provide intuition. 
        \item Inversely, any informal proof provided in the core of the paper should be complemented by formal proofs provided in appendix or supplemental material.
        \item Theorems and Lemmas that the proof relies upon should be properly referenced. 
    \end{itemize}
   
    \item {\bf Experimental Result Reproducibility}
    \item[] Question: Does the paper fully disclose all the information needed to reproduce the main experimental results of the paper to the extent that it affects the main claims and/or conclusions of the paper (regardless of whether the code and data are provided or not)?
    \item[] Answer: \answerYes{} % Replace by \answerYes{}, \answerNo{}, or \answerNA{}.
    \item[] Justification: All experimentation details for training and inference are included in the main and supplementary materials.
    \item[] Guidelines:

    \begin{itemize}
        \item The answer NA means that the paper does not include experiments.
        \item If the paper includes experiments, a No answer to this question will not be perceived well by the reviewers: Making the paper reproducible is important, regardless of whether the code and data are provided or not.
        \item If the contribution is a dataset and/or model, the authors should describe the steps taken to make their results reproducible or verifiable. 
        \item Depending on the contribution, reproducibility can be accomplished in various ways. For example, if the contribution is a novel architecture, describing the architecture fully might suffice, or if the contribution is a specific model and empirical evaluation, it may be necessary to either make it possible for others to replicate the model with the same dataset, or provide access to the model. In general. releasing code and data is often one good way to accomplish this, but reproducibility can also be provided via detailed instructions for how to replicate the results, access to a hosted model (e.g., in the case of a large language model), releasing of a model checkpoint, or other means that are appropriate to the research performed.
        \item While NeurIPS does not require releasing code, the conference does require all submissions to provide some reasonable avenue for reproducibility, which may depend on the nature of the contribution. For example
        \begin{enumerate}
            \item If the contribution is primarily a new algorithm, the paper should make it clear how to reproduce that algorithm.
            \item If the contribution is primarily a new model architecture, the paper should describe the architecture clearly and fully.
            \item If the contribution is a new model (e.g., a large language model), then there should either be a way to access this model for reproducing the results or a way to reproduce the model (e.g., with an open-source dataset or instructions for how to construct the dataset).
            \item We recognize that reproducibility may be tricky in some cases, in which case authors are welcome to describe the particular way they provide for reproducibility. In the case of closed-source models, it may be that access to the model is limited in some way (e.g., to registered users), but it should be possible for other researchers to have some path to reproducing or verifying the results.
        \end{enumerate}
    \end{itemize}

\item {\bf Open access to data and code}
    \item[] Question: Does the paper provide open access to the data and code, with sufficient instructions to faithfully reproduce the main experimental results, as described in supplemental material?
    \item[] Answer: \answerNo{} % Replace by \answerYes{}, \answerNo{}, or \answerNA{}.
    \item[] Justification: We plan to open source the code and datasets pending legal approval.
    \item[] Guidelines:

    \begin{itemize}
        \item The answer NA means that paper does not include experiments requiring code.
        \item Please see the NeurIPS code and data submission guidelines (\url{https://nips.cc/public/guides/CodeSubmissionPolicy}) for more details.
        \item While we encourage the release of code and data, we understand that this might not be possible, so “No” is an acceptable answer. Papers cannot be rejected simply for not including code, unless this is central to the contribution (e.g., for a new open-source benchmark).
        \item The instructions should contain the exact command and environment needed to run to reproduce the results. See the NeurIPS code and data submission guidelines (\url{https://nips.cc/public/guides/CodeSubmissionPolicy}) for more details.
        \item The authors should provide instructions on data access and preparation, including how to access the raw data, preprocessed data, intermediate data, and generated data, etc.
        \item The authors should provide scripts to reproduce all experimental results for the new proposed method and baselines. If only a subset of experiments are reproducible, they should state which ones are omitted from the script and why.
        \item At submission time, to preserve anonymity, the authors should release anonymized versions (if applicable).
        \item Providing as much information as possible in supplemental material (appended to the paper) is recommended, but including URLs to data and code is permitted.
    \end{itemize}

\item {\bf Experimental Setting/Details}
    \item[] Question: Does the paper specify all the training and test details (e.g., data splits, hyperparameters, how they were chosen, type of optimizer, etc.) necessary to understand the results?
    \item[] Answer: \answerYes{} % Replace by \answerYes{}, \answerNo{}, or \answerNA{}.
    \item[] Justification: All experimentation details required for understanding the results are included in the main and supplementary materials.
    \item[] Guidelines:

    \begin{itemize}
        \item The answer NA means that the paper does not include experiments.
        \item The experimental setting should be presented in the core of the paper to a level of detail that is necessary to appreciate the results and make sense of them.
        \item The full details can be provided either with the code, in appendix, or as supplemental material.
    \end{itemize}

\item {\bf Experiment Statistical Significance}
    \item[] Question: Does the paper report error bars suitably and correctly defined or other appropriate information about the statistical significance of the experiments?
    \item[] Answer: \answerYes{}% Replace by \answerYes{}, \answerNo{}, or \answerNA{}.
    \item[] Justification: Yes, mean and standard deviation of the performance metrics are reported across various seed values.
    \item[] Guidelines:

    \begin{itemize}
        \item The answer NA means that the paper does not include experiments.
        \item The authors should answer "Yes" if the results are accompanied by error bars, confidence intervals, or statistical significance tests, at least for the experiments that support the main claims of the paper.
        \item The factors of variability that the error bars are capturing should be clearly stated (for example, train/test split, initialization, random drawing of some parameter, or overall run with given experimental conditions).
        \item The method for calculating the error bars should be explained (closed form formula, call to a library function, bootstrap, etc.)
        \item The assumptions made should be given (e.g., Normally distributed errors).
        \item It should be clear whether the error bar is the standard deviation or the standard error of the mean.
        \item It is OK to report 1-sigma error bars, but one should state it. The authors should preferably report a 2-sigma error bar than state that they have a 96\% CI, if the hypothesis of Normality of errors is not verified.
        \item For asymmetric distributions, the authors should be careful not to show in tables or figures symmetric error bars that would yield results that are out of range (e.g. negative error rates).
        \item If error bars are reported in tables or plots, The authors should explain in the text how they were calculated and reference the corresponding figures or tables in the text.
    \end{itemize}

\item {\bf Experiments Compute Resources}
    \item[] Question: For each experiment, does the paper provide sufficient information on the computer resources (type of compute workers, memory, time of execution) needed to reproduce the experiments?
    \item[] Answer: \answerYes{} % Replace by \answerYes{}, \answerNo{}, or \answerNA{}.
    \item[] Justification: Details of compute used for training and inference are included.
    \item[] Guidelines:

    \begin{itemize}
        \item The answer NA means that the paper does not include experiments.
        \item The paper should indicate the type of compute workers CPU or GPU, internal cluster, or cloud provider, including relevant memory and storage.
        \item The paper should provide the amount of compute required for each of the individual experimental runs as well as estimate the total compute. 
        \item The paper should disclose whether the full research project required more compute than the experiments reported in the paper (e.g., preliminary or failed experiments that didn't make it into the paper). 
    \end{itemize}
\item {\bf Code Of Ethics}
    \item[] Question: Does the research conducted in the paper conform, in every respect, with the NeurIPS Code of Ethics \url{https://neurips.cc/public/EthicsGuidelines}?
    \item[] Answer: \answerYes{} % Replace by \answerYes{}, \answerNo{}, or \answerNA{}.
    \item[] Justification: We conform to NeurIPS code of ethics.
    \begin{itemize}
        \item The answer NA means that the authors have not reviewed the NeurIPS Code of Ethics.
        \item If the authors answer No, they should explain the special circumstances that require a deviation from the Code of Ethics.
        \item The authors should make sure to preserve anonymity (e.g., if there is a special consideration due to laws or regulations in their jurisdiction).
    \end{itemize}

\item {\bf Broader Impacts}
    \item[] Question: Does the paper discuss both potential positive societal impacts and negative societal impacts of the work performed?
    \item[] Answer: \answerYes{} % Replace by \answerYes{}, \answerNo{}, or \answerNA{}.
    \item[] Justification: Yes, discussed in section Appendix \ref{societal-impact}.
    \item[] Guidelines:

    \begin{itemize}
        \item The answer NA means that there is no societal impact of the work performed.
        \item If the authors answer NA or No, they should explain why their work has no societal impact or why the paper does not address societal impact.
        \item Examples of negative societal impacts include potential malicious or unintended uses (e.g., disinformation, generating fake profiles, surveillance), fairness considerations (e.g., deployment of technologies that could make decisions that unfairly impact specific groups), privacy considerations, and security considerations.
        \item The conference expects that many papers will be foundational research and not tied to particular applications, let alone deployments. However, if there is a direct path to any negative applications, the authors should point it out. For example, it is legitimate to point out that an improvement in the quality of generative models could be used to generate deepfakes for disinformation. On the other hand, it is not needed to point out that a generic algorithm for optimizing neural networks could enable people to train models that generate Deepfakes faster.
        \item The authors should consider possible harms that could arise when the technology is being used as intended and functioning correctly, harms that could arise when the technology is being used as intended but gives incorrect results, and harms following from (intentional or unintentional) misuse of the technology.
        \item If there are negative societal impacts, the authors could also discuss possible mitigation strategies (e.g., gated release of models, providing defenses in addition to attacks, mechanisms for monitoring misuse, mechanisms to monitor how a system learns from feedback over time, improving the efficiency and accessibility of ML).
    \end{itemize}
    
\item {\bf Safeguards}
    \item[] Question: Does the paper describe safeguards that have been put in place for responsible release of data or models that have a high risk for misuse (e.g., pretrained language models, image generators, or scraped datasets)?
    \item[] Answer: \answerNA{} % Replace by \answerYes{}, \answerNo{}, or \answerNA{}.
    \item[] Justification: Not applicable since our models do not have high risk of misuse.
    \begin{itemize}
        \item The answer NA means that the paper poses no such risks.
        \item Released models that have a high risk for misuse or dual-use should be released with necessary safeguards to allow for controlled use of the model, for example by requiring that users adhere to usage guidelines or restrictions to access the model or implementing safety filters. 
        \item Datasets that have been scraped from the Internet could pose safety risks. The authors should describe how they avoided releasing unsafe images.
        \item We recognize that providing effective safeguards is challenging, and many papers do not require this, but we encourage authors to take this into account and make a best faith effort.
    \end{itemize}
\item {\bf Licenses for existing assets}
    \item[] Question: Are the creators or original owners of assets (e.g., code, data, models), used in the paper, properly credited and are the license and terms of use explicitly mentioned and properly respected?
    \item[] Answer: \answerYes{} % Replace by \answerYes{}, \answerNo{}, or \answerNA{}.
    \item[] Justification: We follow the license terms for every model and dataset we use.
    \item[] Guidelines:
    \begin{itemize}
        \item The answer NA means that the paper does not use existing assets.
        \item The authors should cite the original paper that produced the code package or dataset.
        \item The authors should state which version of the asset is used and, if possible, include a URL.
        \item The name of the license (e.g., CC-BY 4.0) should be included for each asset.
        \item For scraped data from a particular source (e.g., website), the copyright and terms of service of that source should be provided.
        \item If assets are released, the license, copyright information, and terms of use in the package should be provided. For popular datasets, \url{paperswithcode.com/datasets} has curated licenses for some datasets. Their licensing guide can help determine the license of a dataset.
        \item For existing datasets that are re-packaged, both the original license and the license of the derived asset (if it has changed) should be provided.
        \item If this information is not available online, the authors are encouraged to reach out to the asset's creators.
    \end{itemize}

\item {\bf New Assets}
    \item[] Question: Are new assets introduced in the paper well documented and is the documentation provided alongside the assets?
    \item[] Answer: \answerYes{} % Replace by \answerYes{}, \answerNo{}, or \answerNA{}.
    \item[] Justification: Yes, details of the datasets are provided in the Appendix \ref{visiondataset}.
    \item[] Guidelines:
    \begin{itemize}
        \item The answer NA means that the paper does not release new assets.
        \item Researchers should communicate the details of the dataset/code/model as part of their submissions via structured templates. This includes details about training, license, limitations, etc. 
        \item The paper should discuss whether and how consent was obtained from people whose asset is used.
        \item At submission time, remember to anonymize your assets (if applicable). You can either create an anonymized URL or include an anonymized zip file.
    \end{itemize}
\item {\bf Crowdsourcing and Research with Human Subjects}
    \item[] Question: For crowdsourcing experiments and research with human subjects, does the paper include the full text of instructions given to participants and screenshots, if applicable, as well as details about compensation (if any)? 
    \item[] Answer: \answerNA{} % Replace by \answerYes{}, \answerNo{}, or \answerNA{}.
    \item[] Justification: Not Applicable
    \item[] Guidelines:
    \begin{itemize}
        \item The answer NA means that the paper does not involve crowdsourcing nor research with human subjects.
        \item Including this information in the supplemental material is fine, but if the main contribution of the paper involves human subjects, then as much detail as possible should be included in the main paper. 
        \item According to the NeurIPS Code of Ethics, workers involved in data collection, curation, or other labor should be paid at least the minimum wage in the country of the data collector. 
    \end{itemize}
\item {\bf Institutional Review Board (IRB) Approvals or Equivalent for Research with Human Subjects}
    \item[] Question: Does the paper describe potential risks incurred by study participants, whether such risks were disclosed to the subjects, and whether Institutional Review Board (IRB) approvals (or an equivalent approval/review based on the requirements of your country or institution) were obtained?
    \item[] Answer: \answerNA{} % Replace by \answerYes{}, \answerNo{}, or \answerNA{}.
    \item[] Justification: Not Applicable
    \item[] Guidelines:
    \begin{itemize}
        \item The answer NA means that the paper does not involve crowdsourcing nor research with human subjects.
        \item Depending on the country in which research is conducted, IRB approval (or equivalent) may be required for any human subjects research. If you obtained IRB approval, you should clearly state this in the paper. 
        \item We recognize that the procedures for this may vary significantly between institutions and locations, and we expect authors to adhere to the NeurIPS Code of Ethics and the guidelines for their institution. 
        \item For initial submissions, do not include any information that would break anonymity (if applicable), such as the institution conducting the review.
    \end{itemize}
\end{enumerate}